\documentclass[9pt,twocolumn]{extarticle}
\usepackage{modernstyle}

\usepackage[round]{natbib}

\usepackage{hyperref}
\usepackage{url}
\usepackage{adjustbox}

\usepackage[utf8]{inputenc}
\usepackage{booktabs}
\usepackage{nicefrac}  
\usepackage{microtype} 
\usepackage{xcolor}    
\usepackage{multirow}
\usepackage{colortbl}
\usepackage{listings}
\usepackage{subcaption}
\usepackage{float}
\usepackage{CJKutf8}
\usepackage{amsmath}
\usepackage{amsfonts}
\definecolor{headercolor}{gray}{0.9}

\usepackage{fancyhdr}
\fancypagestyle{firstpage}{%
	\fancyhf{}
	\setlength{\headsep}{0.3in}
	\cfoot{\thepage}
}
\fancypagestyle{otherpages}{%
	\fancyhf{}
	\setlength{\headsep}{0.3in}
	\rhead{XAI-Units}
	\lhead{J. R. Lee, S. Emami, M. Hollins, T. Wong, C. Villalobos Sánchez, F. Toni, D. Zhang, A. Dejl}
	\cfoot{\thepage}
}

\title{XAI-Units: Benchmarking Explainability Methods with Unit Tests}

\author{
	Jun Rui Lee\\
	\normalsize Department of Computing\\
	\normalsize Imperial College London\\
	\texttt{junruilee@yahoo.com}
	\AuthorAnd
	Sadegh Emami\\
	\normalsize Department of Computing\\
	\normalsize Imperial College London\\
	\texttt{sadegh@emami.net}
	\AuthorAnd
	Michael David Hollins\\
	\normalsize Department of Computing\\
	\normalsize Imperial College London\\
	\texttt{michaelhollins@hotmail.co.uk}
	\AuthorAND
	Timothy C. H. Wong\\
	\normalsize Department of Computing\\
	\normalsize Imperial College London\\
	\texttt{timchwong1919@gmail.com}
	\AuthorAnd
	Carlos Ignacio Villalobos Sánchez\\
	\normalsize Department of Computing\\
	\normalsize Imperial College London\\
	\texttt{carlosignaciovillalobos99@gmail.com}
	\AuthorAnd
	Francesca Toni\\
	\normalsize Department of Computing\\
	\normalsize Imperial College London\\
	\texttt{ft@imperial.ac.uk}
	\AuthorAND
	Dekai Zhang\\
	\normalsize Department of Computing\\
	\normalsize Imperial College London\\
	\texttt{dz819@imperial.ac.uk}
	\AuthorAnd
	Adam Dejl\\
	\normalsize Department of Computing\\
	\normalsize Imperial College London\\
	\texttt{adam.dejl18@imperial.ac.uk}
}

\begin{document}
	\pagestyle{otherpages}
	\maketitle
	
	\begin{abstract}
		Feature attribution (FA) methods are widely used in explainable AI (XAI) to help users understand how the inputs of a machine learning model contribute to its outputs. However, different FA models often provide disagreeing importance scores for the same model. In the absence of ground truth or in-depth knowledge about the inner workings of the model, it is often difficult to meaningfully determine which of the different FA methods produce more suitable explanations in different contexts. As a step towards addressing this issue, we introduce the open-source \textsc{XAI-Units} benchmark, specifically designed to evaluate FA methods against diverse types of model behaviours, such as feature interactions, cancellations, and discontinuous outputs.\footnote{The benchmark package is available at \href{https://github.com/XAI-Units/xaiunits}{https://github.com/XAI-Units/xaiunits}} 
		Our benchmark provides a set of paired datasets and models with known internal mechanisms, establishing clear expectations for desirable attribution scores. Accompanied by a suite of built-in evaluation metrics, \textsc{XAI-Units} streamlines systematic experimentation and reveals how FA methods perform against distinct, atomic kinds of model reasoning, similar to unit tests in software engineering. Crucially, by using procedurally generated models tied to synthetic datasets, we pave the way towards an objective and reliable comparison of FA methods.
	\end{abstract}
	
	\section{Introduction}\label{sec:introduction}
	
	As artificial intelligence (AI) and machine learning (ML) techniques are increasingly embraced, the importance of interpreting these models through explainable AI (XAI) techniques also grows. By improving users' understanding of the logic behind AI models, XAI offers benefits in various settings including increasing social acceptance and trust, meeting legal obligations, detecting and removing bias, debugging unanticipated behaviour and enhancing AI safety.
	
	\begin{figure*}[htb]
		\centering
		\includegraphics[width=0.75\linewidth]{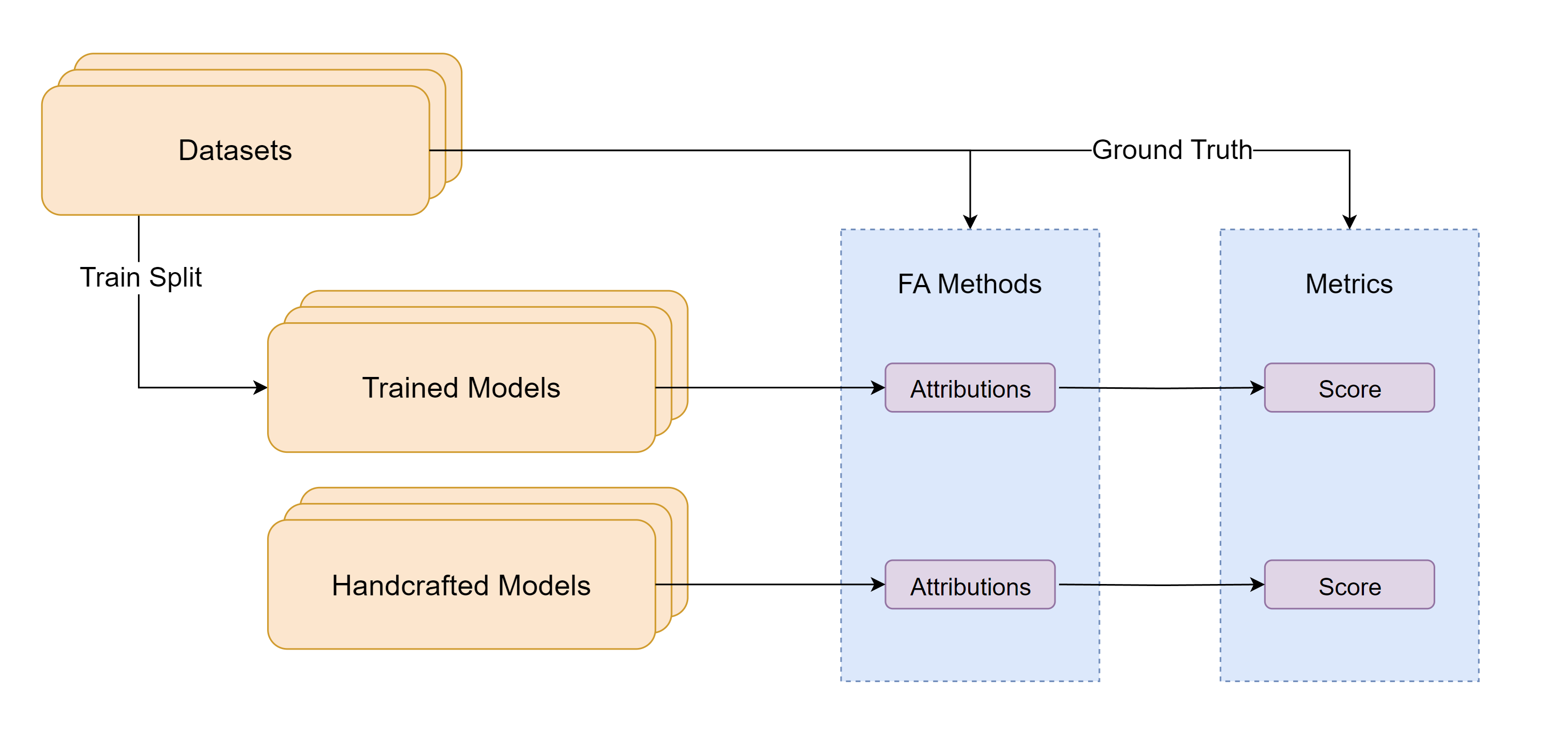}
		\caption{Overview of the \textsc{XAI-Units} benchmark. The benchmark provides a set of datasets and models with controlled mechanisms and behaviour. This enables us to evaluate the attributions produced by various FA methods using various metrics, often taking into account the ground-truth expectations associated with the given dataset and model.}
		\label{fig:package_workflow}
	\end{figure*}
	
	Feature attribution (FA) methods are a branch of XAI focused on quantifying the effects of input features on model outputs. Common methods include perturbation-based ones such as \texttt{LIME}~\citep{ribeiro2016why} and \texttt{SHAP}~\citep{lundberg2017unified}, or gradient-based ones such as \texttt{DeepLIFT}~\citep{shrikumar2019learning} and \texttt{Integrated} \texttt{Gradients}~\citep{sundararajan2017axiomatic}. These approaches reduce the complexity of a model's mathematical logic into a set of numerical scores which quantify the importance of each feature.
	
	However, as the number of proposed FA methods has increased, practitioners have encountered confusing situations where these methods contradict each other~\citep{roy2022}, also known as the \textit{disagreement problem}~\citep{krishna2022disagreement}. This evidently undermines the motivation behind FA methods, which is to disambiguate the reasoning process of an ML model. In response, a variety of metrics have been proposed for evaluating FA methods~\citep{zhou2021evaluating, nauta2023}. However, due to the difficulty of establishing what constitutes a ``better'' explanation in various scenarios, these metrics are often merely heuristic and may not accurately rank the performance of FA methods.
	
	With the proliferation of different FA methods and evaluation metrics, a practical need arose to simplify the burgeoning complexity of XAI analysis. Therefore, various XAI toolkits were developed to streamline the comparison between datasets, models, FA methods and metrics~\citep{le2023benchmarking, liu2021synthetic, hedstrom2023quantus, agarwal2024openxai}. While this has helped the research process, what remains elusive is establishing the conditions under which FA methods reliably capture the internal ``reasoning" process of models.
	
	This challenge motivates our \textsc{XAI-Units} package (Figure~\ref{fig:package_workflow}), which allows the user to assess how various FA methods perform against expected, atomic units of model behaviour. This highlights the respective strength and limitations of respective FA methods, contributing to greater transparency for users seeking to understand and trust model explanations. By using deterministic models fully aligned with our synthetic datasets and avoiding complex datasets with unclear mechanisms, we are able to provide ground truths facilitating better understanding and evaluation of FA methods. Therefore, while \textsc{XAI-Units} focuses on synthetic test cases, this choice is deliberate, enabling us to systematically evaluate FA methods against predictable and distinctive units of model behaviour, which is challenging to achieve with real-world datasets. Although synthetic data simplifies complex real-world scenarios, many of the situations we test, such as feature interaction effects, mirror challenges observed in domains such as healthcare and finance. For those seeking benchmarks on real-world datasets, we refer to complementary efforts in prior work outlined towards the end of Section~\ref{sec:RelatedWork}. Moreover, we envision future extensions of our framework to include semi-synthetic or real-world applications.
	
	Our aim with this paper is not to provide definitive reasons as to why some FA methods perform badly in certain settings. Rather, we provide a package which enables researchers to easily find under what conditions particular FA methods struggle, with the hope that this will prompt further exploration into the reasons behind it. To summarise our contributions:
	\begin{enumerate}
		\item We provide a benchmark for evaluating FA methods, enabling developers to verify that XAI techniques meet their design specifications and ensuring accountability for correct implementation. Our key approach is to either procedurally generate (``handcraft") or engineer a collection of neural network models to replicate specific types of testable behaviours.
		\item We create corresponding synthetic data generators that are paired with each of our models to enable a controlled environment for evaluation.
		\item We implement the entire benchmarking pipeline within the open source Python library \textsc{XAI-Units}, which is fully extensible to support custom evaluation metrics and FA methods.
		\item We apply our benchmark to common FA methods, testing their strengths and weaknesses on specific model behaviour. Using this approach, we identify an implementation discrepancy in a popular FA library.
	\end{enumerate}
	
	The rest of this paper is organised as follows. In Section \ref{sec:RelatedWork}, we describe the related work on evaluating and benchmarking feature attribution methods. Section \ref{sec:package} presents the \textsc{XAI-Units} benchmark, and the datasets and models included in it. In Section \ref{sec:example_use_case}, we report results from applying \textsc{XAI-Units} to common FA methods. Finally, in Section \ref{sec:conclusion}, we conclude with an overall discussion of the work.

	\section{Related work} \label{sec:RelatedWork}
	
	\begin{table*}[htb]
		\caption{Comparison between \textsc{XAI-Units} and existing toolkits.}
		\centering
		\label{tab:prior_work}
		\begin{tabular}{@{}lccccc@{}}
			\toprule
			Toolkit & Real-World  & Synthetic & Ground-Truth & Extensible & Model \\
			& Datasets &  Datasets & Available &  & Behaviour-focused \\         
			\midrule
			\textsc{XAI-Units} & No & Yes          & Yes &  Yes & Yes    \\
			\textsc{OpenXAI}~\citep{agarwal2024openxai} & Yes & Yes    & No   & Yes  & No     \\
			\textsc{Quantus}~\citep{hedstrom2023quantus} & No & No    & No   & Yes   & No    \\
			$\mathcal{M}^4$~\citep{li2023mathcal} & Yes & Yes    & Partial   & Yes  & No     \\
			\textsc{GraphXAI}~\citep{agarwal2023evaluating} & Yes & Yes & Yes & No  & No \\
			\textsc{XAI-Bench}~\citep{liu2021synthetic} & No & Yes & Yes & No & No \\
			\textsc{BAM}~\citep{yang2019benchmarking} & No & Yes & No & No & No\\
			\bottomrule
		\end{tabular}
	\end{table*}
	
	\subsection{Evaluation of feature attribution methods}
	
	Feature attributions (FA) are a category of XAI methods that calculate attribution scores for all input features for a given model~\citep{zhou2022feature}. These scores help to delineate each feature's impact and importance on the model outcome. Numerous FA methods have been introduced in the literature~\citep{ribeiro2016why, lundberg2017unified, dabkowski2017real, ramaswamy2020ablation, shrikumar2019learning, bach2015pixel}, and can be broadly grouped into two main categories depending on whether they are based on gradients or perturbations~\citep{ancona2018better} (see \citet{speith2022} for a more detailed review of XAI method taxonomies). Yet the differing approaches may lead to different attribution scores, which is commonly referred to as a \textit{disagreement problem}~\citep{krishna2022disagreement}. This issue has been studied in detail for two widely used explainability methods, \texttt{LIME} and \texttt{SHAP}~\citep{roy2022}. Thus, to assess the quality of the attribution scores across FA methods, there is a clear need for their reliable evaluation.
	
	Several evaluation metrics for FA methods have been proposed following questions around the effectiveness and consistency across different FA methods~\citep{krishna2022disagreement, Bilodeau_2024}. By evaluating FA methods, researchers and users may get a better idea as to which method performs better in a particular case. Accordingly, different evaluation metrics are used to test FA methods against certain desirable properties~\citep{lundberg2017unified, sundararajan2017axiomatic}. For instance, for a given FA method, the \textit{explanation infidelity} metric aims to capture its faithfulness, whilst the \textit{explanation sensitivity} (or \textit{max-sensitivity}) metric measures its robustness~\citep{yeh2019infidelity, alvarezmelis2018robustness}. That said, these metrics typically only offer a rough indication of an FA method's performance relative to other competing methods: in most cases, there is no unambiguous metric indicating the absolute quality of any FA method due to the absence of ground truth attributions. Some FA evaluation methods are specific to certain modalities, such as the metrics based on 2x2 image grids proposed by \citet{Fresz2024} for evaluating explanations in computer vision.
	
	\subsection{Synthetic approaches}
	
	Synthetic datasets are utilised in many of the existing benchmarks for FA methods because ground truth attributions can be derived from them~\citep{zhou2022feature, arras2022clevr, agarwal2023evaluating, zhang2024attributionlab}. Different benchmarks generate their synthetic datasets differently. XAI-Bench constructs their tabular synthetic datasets via mimicking common statistical distributions~\citep{liu2021synthetic} and the ground truth attributions are derivable from the corresponding distribution. Similarly, the synthetic data generator from \textsc{OpenXAI} is based on sampling clusters from Gaussian distributions, which, given the most accurate possible model, has been proven to facilitate the computation of ground-truth attributions for each cluster~\citep{agarwal2024openxai}. However, there is no guarantee that these ``ground-truth'' attributions will be aligned with imperfect models that result from the used training procedures.
	The tabular benchmarks in \textsc{XAI-Units} are also based on synthetic data sampled from statistical distributions, but in contrast to the packages mentioned above, each dataset is paired with a handcrafted model with perfect accuracy. Additionally, \textsc{XAI-Units} datasets and models are focused on atomic model behaviours rather than generic statistical distributions.
	
	Handcrafted neural network models are rare in most existing benchmarks because in real-world applications, models are trained to fit the observed data rather than predefined. However, as our goal is not to replicate real-world scenarios but to create a controlled environment to test FA methods, we pair synthetic datasets with handcrafted models. As argued in \citet{breiman2001statistical}, this is motivated by the \textit{Rashomon effect}, where the ground-truth attributions derived from synthetic datasets are trustworthy and effective only when they are paired with handcrafted models during the evaluation of FA methods. With our focus on using ground-truth attributions to evaluate FA methods, our approach is related to the \textit{Synthetic Explainable Classifier} generators of \citet{guidotti2021evaluating}. However, our focus is on how FA methods perform against particular, controlled, distinct kinds of model reasoning. Therefore, our approach offers several potential benefits to the XAI community, such as allowing researchers to test how well new FA methods handle particular types of feature interaction. 
	
	\begin{table*}[htb]
		\caption{Summary of available synthetic data generators in \textsc{XAI-Units}.}
		\centering
		\label{tab:available_datasets}
		% \resizebox{\linewidth}{!}{%
			\begin{tabular}{@{}lcccc@{}}
				\toprule
				Data Generator    &   Datatype   & Feature type(s)        & Ground Truth  &  Default Metric \\ 
				
				\midrule
				Weighted Continuous & Tabular &Continuous          & Available &     MSE      \\
				Conflicting Features & Tabular &Continuous, Categorical     & Available   & MSE       \\
				Pertinent Negatives & Tabular &Continuous, Categorical    & Available   &    MSE     \\
				Feature Interaction & Tabular &Continuous, Categorical    & Available   &     MSE    \\
				Uncertainty & Tabular &Continuous          & Mask     &    Mask Error   \\
				Shattered Gradient  & Tabular &Continuous        & Unavailable   &    \texttt{SensitivityMax}     \\
				Boolean Formula & Tabular &Categorical         & Unavailable    &    \texttt{Infidelity}     \\
				\hspace{0.5cm} Boolean AND & Tabular  & Categorical & Available & MSE \\
				\hspace{0.5cm} Boolean OR & Tabular & Categorical & Available & MSE \\
				Balanced Image & Images &          & Mask    &    Mask Proportion Image     \\
				Imbalanced Image & Images &          & Mask    &    Mask Proportion Image     \\
				Trigger Injection & Text &         & Mask    &    Mask Proportion Text   \\
				\bottomrule
			\end{tabular}
			% }
	\end{table*}
	
	\subsection{Toolkits for evaluating feature attribution methods}
	
	Toolkits for FA methods have been developed for the easy application of evaluation metrics and benchmarking across different FA methods. The benchmarks offered by existing toolkits are generally constructed based on either real-world~\citep{zhang2023xai, huang2023safari, lin2020see, cui-etal-2022-expmrc} or synthetic datasets~\citep{liu2021synthetic, mamalakis2022neural, yang2019benchmarking}, or a blend of both~\citep{agarwal2024openxai, li2023mathcal, agarwal2023evaluating}. According to a recent survey on existing toolkits, \textsc{OpenXAI} and \textsc{Quantus} are two of the most popular options~\citep{le2023benchmarking}. \textsc{OpenXAI}~\citep{agarwal2024openxai} constructs its benchmark by applying FA methods to trained models on real-world datasets. A synthetic data generator that generates multiple clusters of normally-distributed data is also available for benchmarking. Moreover, it offers an open-source end-to-end \texttt{Pipeline} for implementing FA methods and evaluating them. Meanwhile, \textsc{Quantus}~\citep{hedstrom2023quantus} is a Python package that gathers a diverse pool of over 30 different built-in evaluation metrics and is extendable to custom evaluation metrics. Although \textsc{Quantus} does not contain any pre-loaded datasets, the user can load in their own data and use the implemented metrics to evaluate FA methods and other XAI methods with respect to various properties. 
	
	Sharing commonalities with \textsc{OpenXAI} and \textsc{Quantus}, \textsc{XAI-Units} provides a complete \texttt{Pipeline} for benchmarking FA methods and is able to support custom methods and metrics during evaluation. However, our work distinguishes itself from the existing toolkits by enabling the evaluation of FA methods in a more controlled setting. Enabled by the usage of procedurally generated handcrafted models with known internal mechanisms, each dataset and model pair is analogous to a unit test that isolates a single type of input behaviour to test for. This effectively circumvents the \textit{blame problem}~\citep{rahnama2023blame}, as our toolkit eliminates the ambiguity in deciding whether a FA method's poor performance is driven by the method itself or the model behaviour. Moreover, \textsc{XAI-Units} is compatible with a broad range of data modalities and model architectures. In particular, the toolkit incorporates multilayer perceptrons (MLPs), convolutional neural networks (CNNs), vision transformers (ViTs) and large language models (LLMs), while also supporting diverse modalities including tabular data, images and text.

	\section{Package overview} \label{sec:package}
	Here, we describe \textsc{XAI-Units}' core components, as outlined in Figure~\ref{fig:package_workflow}. We focus on the datasets and models included in the benchmark, followed by the tested FA methods and the used metrics.
	
	\subsection{Datasets and models} 
	\label{sec:datasets_and_models}
	
	\textsc{XAI-Units} contains seven tabular, two image, and one text synthetic data generators, summarised in Table~\ref{tab:available_datasets}. Each tabular dataset generator is paired with a handcrafted neural network model whose logic is set out formally in Appendix \ref{sec:nn_diagrams_appendix}. Motivating each of these pairs are distinct, simple units of behaviour which some FA methods may struggle with. Clearly, this set of pairs is not exhaustive, but by making \textsc{XAI-Units} easily extensible, we encourage other researchers to %devise and 
	add their own ``unit tests''. Apart from the handcrafted models, the benchmark also provides analogous trained models for comparison.
	
	\paragraph{Weighted Continuous} This neural network returns a weighted sum of input features. It serves as a simple baseline for evaluating FA methods and a springboard for developing the more complicated models. It implements a linear function using a two-layer MLP network with \texttt{ReLU} activations in the hidden layer.
	
	\paragraph{Conflicting Features} This dataset-model pair introduces categorical ``cancellation'' features which cancel or negate the impact of continuous features. This tests how well FA methods handle cases where there are conflicts between features. Aim to surface such conflicts has previously informed the design of several FA methods, including DeepLIFT RevealCancel~\citep{shrikumar2017just} and CAFE~\citep{dejl2025conflicts}. As an example of a conflict between features, consider a healthcare AI system predicting a patient’s risk of death based on vital signs and previous treatments. When faced with a normal temperature reading, the system may typically predict a lower overall risk, but this line of reasoning may be void when the patient was recently administered an antipyretic drug.

	\paragraph{Pertinent Negatives} This dataset-model pair captures scenarios where an output meaningfully depends on the zero value of a (pertinent negative) feature. Such feature behaviour may be present in models that predict the likelihood that a patient is suffering from a severe heart condition given the heart rate. As the $0$ heart rate is meaningful for the prediction (indicating asystole), FA methods should ideally return non-zero attributions for this feature. However, this may trouble FA methods as the $0$ feature value may lead to a $0$ attribution score being returned.

	\paragraph{Shattered Gradients} The logic of shattered gradients is that minor input changes that have negligible impact on the model output can lead to significant changes in attribution scores. An example of this is the point of gradient discontinuity in the following function $\text{ReLU}(x-100)$. Any infinitesimal positive perturbation around the discontinuity will still result in similar output however, gradients and thus attribution would change drastically relative to the magnitude of the perturbation.

	\paragraph{Categorical Feature Interaction} This dataset-model pair is based on interactions between categorical and continuous features. Each continuous feature's weight varies depending on the associated categorical feature's value — either $0$ or $1$. Specifically, for any given pair consisting of a continuous feature and a categorical feature, the weights are defined as $(w^{(1)}, w^{(2)})$. If the categorical feature's value is $0$, the weight applied to the continuous feature is $w^{(1)}$. Conversely, if the categorical feature's value is $1$, the weight becomes $w^{(2)}$. 
	An example of the interacting features logic is in predicting a client's credit score; the importance of their salary may depend on (or interact with) whether they are ``old'' or ``young''.

	\paragraph{Uncertainty Model} This model-dataset pair captures when a subset of input features is irrelevant for probabilistic class prediction. In the case of a classifier that is given redundant inputs (with no impact on the prediction), one would expect a perfect FA method to not assign attribution scores to these inputs.
	
	The model is composed of a linear transformation followed by a \texttt{softmax} activation layer. In this model, some input features are irrelevant to output class prediction and so are designated as \texttt{common}. These \texttt{common} features simply add a constant term to all output class logits equally. Thus, as the \texttt{softmax} layer is translation invariant $f(x+b) = f(x)$ where $x,b$ are vectors, \texttt{common} features have no impact on the output class prediction. Thus the default evaluation metric is Mask Error (defined in Appendix~\ref{sec:appendix_uncertainty}) which penalises explanations that give larger attributions to the \texttt{common} features.
	
	\paragraph{Boolean Formula} The final tabular datasets and models provided are Boolean formulae. Using this framework, the user may implement neural networks that replicate the logic of any arbitrary Boolean formula that is made up of the `AND', `OR', and `NOT' connectives. The dataset for any such formula consists of permutations of truth values for the propositional atoms. Each propositional atom is represented as $+1$ if its value is \texttt{True}, or as $-1$ if its value is \texttt{False}. The significance of this model-dataset pair is given by the common existence of sufficient/necessary conditions for a particular prediction and the general utility of logical rules in reasoning.
	
	\begin{figure*}[htb]
		\begin{minipage}{0.18\linewidth}
			\includegraphics[width=\linewidth]{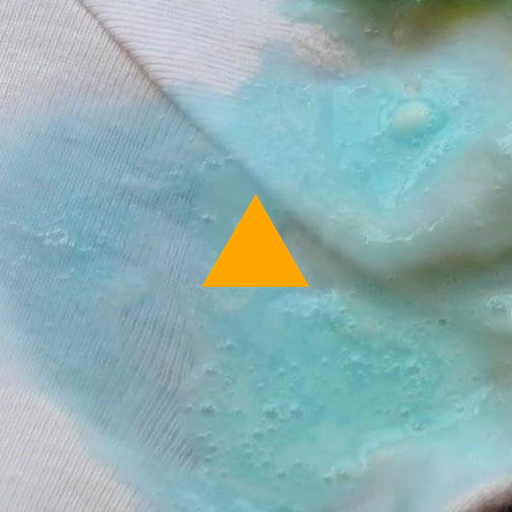}
			\subcaption{Original}
		\end{minipage}\hfill
		\begin{minipage}{0.18\linewidth}
			\includegraphics[width=\linewidth]{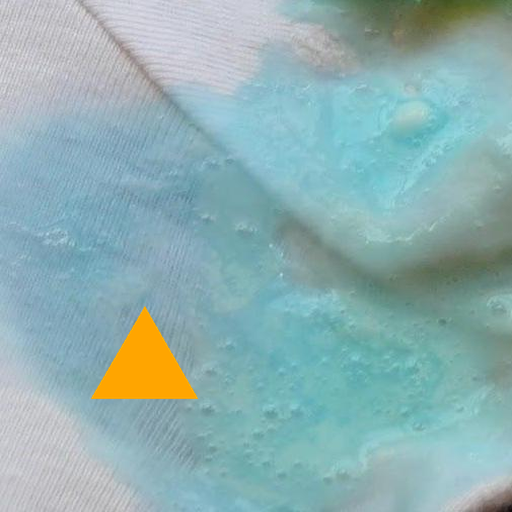}
			\subcaption{Position}
		\end{minipage}\hfill
		\begin{minipage}{0.18\linewidth}
			\includegraphics[width=\linewidth]{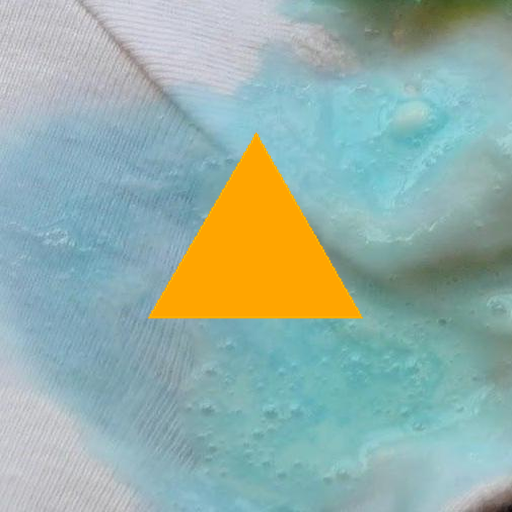}
			\subcaption{Size}
		\end{minipage}\hfill
		\begin{minipage}{0.18\linewidth}
			\includegraphics[width=\linewidth]{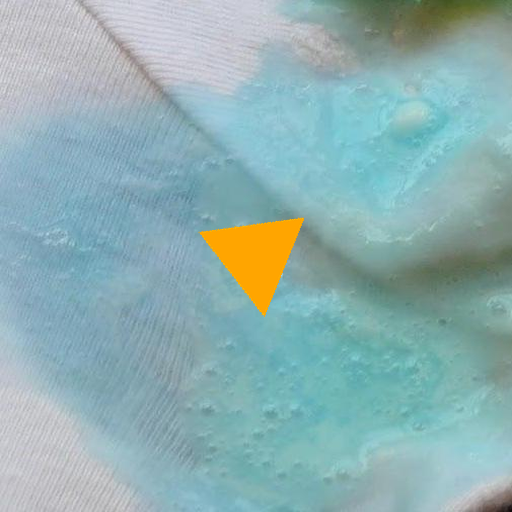} 
			\subcaption{Rotation}
		\end{minipage}\hfill
		\begin{minipage}{0.18\linewidth}
			\includegraphics[width=\linewidth]{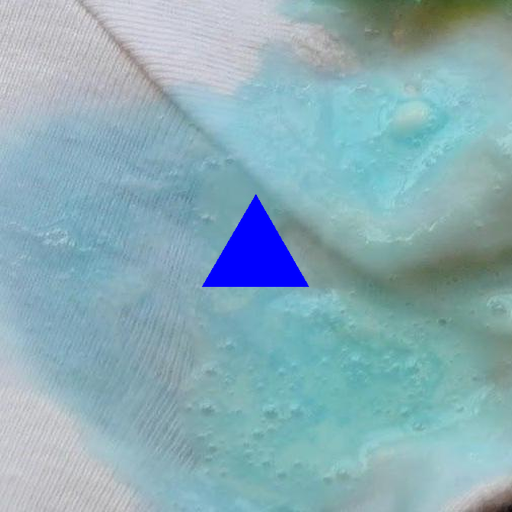}
			\subcaption{Colour}
		\end{minipage}\hfill
		\caption{Variations of a foreground-background combination from the \texttt{BalancedImageDataset}.}
		\label{fig:ImageDatasetExamples}
	\end{figure*}
	
	\paragraph{Image datasets} In addition to the tabular datasets, we also provide image datasets which are designed to overlay various foregrounds (geometric shapes or images of dinosaurs) onto diverse, textured backgrounds.\footnote{Dinosaur image files were sourced from Wikimedia Commons, the free media repository, and licensed under the Creative Commons license CC BY-SA \citep{wiki2024}. Textured backgrounds were taken from the Describable Textures Dataset (DTD), available to the computer vision community for research purposes \citep{cimpoi14describing2014}.} This setup is crucial for creating scenarios that closely mimic real-world conditions where objects of interest (foregrounds) appear against varying scenes (backgrounds). A good FA method applied to a well-trained model might assign higher attribute scores to pixels in the foreground rather than the irrelevant background. 
	
	For a given sample (combining a foreground and background), each image can be customised in terms of the position, size, rotation, and colour of the foreground objects (see Figure \ref{fig:ImageDatasetExamples}). These attributes can be set to fixed values or varied randomly, thereby introducing necessary variations that challenge the robustness of image recognition models. In addition, the library has the ability to generate both balanced and imbalanced datasets. In a balanced dataset, every combination of background, foreground, and colour appears an equal number of times, which is ideal for basic model training where equal representation ensures unbiased learning. The imbalanced dataset, conversely, simulates real-world scenarios where certain objects might appear more frequently with specific backgrounds, causing the model to focus more on the background with the attribution scores changing accordingly.
	
	\paragraph{Text datasets} Given the rapid adoption and evolution of LLMs, our benchmark also includes a dedicated text dataset and corresponding LLMs, enabling practitioners to experiment with the application of FA methods on these models. However, using standard LLM and generic prompts would prohibit us from objectively comparing the attribution scores across the different FA methods, as we are unable to objectively identify which tokens are necessary for the next token generation.  
	
	In line with the philosophy of our package, we provide a ``unit test" for FA method on LLMs called ``Trigger Injection", drawing inspiration from \citet{saha2019hiddentriggerbackdoorattacks} and \citet{yan2024backdooringinstructiontunedlargelanguage}. ``Trigger Injection" has two components, a dataset of prompts with Trigger Words embedded within, and a fine-tuned \texttt{Llama-3.2-1B-Instruct} called \texttt{TriggerLLM}. 
	
	\texttt{TriggerLLM} has been fine-tuned such that, in the presence of the Trigger Word in a prompt, the LLM will respond by only generating the Trigger Response Token. Otherwise, in the absence of the Trigger Word in a prompt, \texttt{TriggerLLM} will generate responses as per usual. Training the model to respond to the trigger word provides clear expectations of the model's behaviour, thus enabling the direct comparison of attributions scores. We report further details about our fine-tuning process for \texttt{TriggerLLM} in Appendix \ref{sec:text_model}. Apart from being a useful model of data backdoor attacks, our ``Trigger Injection" unit test can also serve as a way to isolate the effects of specific instructions in the prompt on the model behaviour.
	
	\subsection{FA methods and evaluation metrics}
	
	Having covered the dataset and model components, here we briefly review the integrated FA methods and metrics. To ensure compatibility with the existing ecosystem, the \textsc{XAI-Units} package natively supports running \textsc{Captum}~\citep{kokhlikyan2020captum} FA methods (e.g. \texttt{DeepLIFT}, \texttt{ShapleyValueSampling}~\citep{castro2009polynomial}) and also contains a wrapper class for running custom FA methods. The package also supports \textsc{Captum}'s official attribution wrappers for LLMs. Similarly, our package supports the evaluation metrics from \textsc{Captum} (\texttt{Infidelity}, \texttt{SensitivityMax}), as well as providing a wrapper class for running custom metrics. Table \ref{tab:available_datasets} displays the default evaluation metric used for each dataset in the package: in cases where a ground truth attribution can be determined, mean squared error (or a variant thereof) is used as the default metric. Our ground truth attributions were derived by ablating inputs to a baseline reference value (in most cases $0$). The full definitions and the associated details are provided in Appendix~\ref{sec:nn_diagrams_appendix}.

	\section{Benchmarking analysis} \label{sec:example_use_case}
	
	We now demonstrate how experiments run with the \textsc{XAI-Units} package can provide novel insights for evaluating FA methods. The code and instructions to reproduce all results in this section are provided in the supplementary materials.
	
	\subsection{Tabular dataset experiments} \label{sec:tabular_experiments}
	To start with, we tested the performance of several common FA methods on the tabular datasets and handcrafted models outlined in Section~\ref{sec:datasets_and_models}. For comparison, we ran identical experiments for both our handcrafted model and trained models.
	
	\begin{table*}
		\centering
		\caption{Tabular Dataset Results. $\downarrow / \uparrow$ indicates a low / high score is better.}
		\label{tab:tabular_results}
		\resizebox{0.75\textwidth}{!}{%
			\centering
			\begin{tabular}{llcccc}
				\toprule
				&  & \multicolumn{4}{c}{\textbf{Default Metric}$^{\downarrow}$} \\
				& & \textbf{Weighted Fts}\textsuperscript{\textit{a}} & \textbf{Conflicting}\textsuperscript{\textit{b}} & \textbf{Interacting}\textsuperscript{\textit{c}} & \textbf{Uncertainty}\textsuperscript{\textit{d}} \\
				\cline{1-6}
				\multicolumn{2}{l}{\texttt{DeepLIFT}} \\ 
				& Handcrafted & 0.000 ± 0.000 & 0.175 ± 0.040 & 0.000 ± 0.000 & 62.915 ± 80.537 \\
				& Trained & 0.003 ± 0.002 & 0.061 ± 0.023 & 0.089 ± 0.076 & 0.002 ± 0.000 \\
				\cline{1-6}
				\multicolumn{2}{l}{\texttt{InputXGradient}} \\ 
				& Handcrafted & 0.000 ± 0.000 & 0.175 ± 0.040 & 0.000 ± 0.000 & 0.000 ± 0.000 \\
				& Trained & 0.006 ± 0.003 & 0.108 ± 0.026 & 0.124 ± 0.113 & 0.005 ± 0.001 \\
				\cline{1-6}
				\multicolumn{2}{l}{\texttt{IntegratedGradients}} \\ 
				& Handcrafted & 0.000 ± 0.000 & 0.175 ± 0.040 & 0.000 ± 0.000 & 0.000 ± 0.000 \\
				& Trained & 0.003 ± 0.002 & 0.060 ± 0.023 & 0.084 ± 0.070 & 0.006 ± 0.002 \\
				\cline{1-6}
				\multicolumn{2}{l}{\texttt{KernelSHAP}} \\ 
				& Handcrafted & 0.000 ± 0.000 & 0.470 ± 0.116 & 0.128 ± 0.108 & 0.000 ± 0.000 \\
				& Trained & 0.002 ± 0.001 & 0.574 ± 0.113 & 0.131 ± 0.105 & 0.007 ± 0.002 \\
				\cline{1-6}
				\multicolumn{2}{l}{\texttt{ShapleyValueSampling}} \\ 
				& Handcrafted & 0.000 ± 0.000 & 0.055 ± 0.013 & 0.073 ± 0.067 & 0.000 ± 0.000 \\
				& Trained & 0.001 ± 0.000 & 0.068 ± 0.022 & 0.074 ± 0.067 & 0.001 ± 0.000 \\
				\cline{1-6}
				\multicolumn{2}{l}{\texttt{LIME} (Linear)} \\ 
				& Handcrafted & 0.000 ± 0.000 & 0.179 ± 0.040 & 0.099 ± 0.082 & 0.000 ± 0.000 \\
				& Trained & 0.001 ± 0.001 & 0.253 ± 0.059 & 0.104 ± 0.084 & 0.004 ± 0.001 \\
				\cline{1-6}
				\multicolumn{2}{l}{\texttt{LIME} (Lasso)} \\ 
				& Handcrafted & 0.005 ± 0.001 & 0.092 ± 0.022 & 0.090 ± 0.075 & 0.000 ± 0.000 \\
				& Trained & 0.006 ± 0.001 & 0.113 ± 0.030 & 0.093 ± 0.075 & 0.001 ± 0.000 \\
				\hline
				\multicolumn{2}{l}{\cellcolor{headercolor}Model Performance\textsuperscript{\textit{e}}} &\cellcolor{headercolor} &\cellcolor{headercolor} &\cellcolor{headercolor} &\cellcolor{headercolor} \\
				\cellcolor{headercolor}
				& \cellcolor{headercolor} Handcrafted & \cellcolor{headercolor} 0.000 ± 0.000$^{\downarrow}$ & \cellcolor{headercolor} 0.000 ± 0.000$^{\downarrow}$ & \cellcolor{headercolor} 0.000 ± 0.000$^{\downarrow}$ & \cellcolor{headercolor} 1.000 ± 0.000$^{\uparrow}$ \\
				\cellcolor{headercolor}
				& \cellcolor{headercolor} Trained & \cellcolor{headercolor} 0.006 ± 0.004$^{\downarrow}$ & \cellcolor{headercolor} 0.160 ± 0.081$^{\downarrow}$ & \cellcolor{headercolor} 0.032 ± 0.026$^{\downarrow}$ & \cellcolor{headercolor} 0.944 ± 0.012$^{\uparrow}$ \\
				\hline
				& & \textbf{Shattered Grad}\textsuperscript{\textit{f}} & \textbf{Pertinent Neg}\textsuperscript{\textit{g}} & \textbf{Bool AND}\textsuperscript{\textit{h}} & \textbf{Bool OR}\textsuperscript{\textit{i}} \\
				\hline
				\multicolumn{2}{l}{\texttt{DeepLIFT}} \\ 
				& Handcrafted & 1.896 ± 0.147 & 0.000 ± 0.000 & 0.000 ± 0.000 & 0.000 ± 0.000 \\
				& Trained & 17.510 ± 8.520 & 0.351 ± 0.520 & 0.045 ± 0.034 & 0.033 ± 0.025 \\
				\cline{1-6}
				\multicolumn{2}{l}{\texttt{InputXGradient}} \\ 
				& Handcrafted & 1.896 ± 0.147 & 11.899 ± 4.649 & 0.094 ± 0.002 & 3.287 ± 0.005 \\
				& Trained & 96.579 ± 49.496 & 0.953 ± 1.419 & 0.066 ± 0.002 & 0.066 ± 0.004 \\
				\cline{1-6}
				\multicolumn{2}{l}{\texttt{IntegratedGradients}} \\ 
				& Handcrafted & 1.896 ± 0.147 & 0.000 ± 0.000 & 0.000 ± 0.000 & 0.000 ± 0.000 \\
				& Trained & 17.943 ± 8.438 & 0.345 ± 0.514 & 0.044 ± 0.035 & 0.034 ± 0.025 \\
				\cline{1-6}
				\multicolumn{2}{l}{\texttt{KernelSHAP}} \\ 
				& Handcrafted & 2.089 ± 0.364 & 0.000 ± 0.000 & 0.355 ± 0.008 & 0.352 ± 0.004 \\
				& Trained & 4.652 ± 1.020 & 0.301 ± 0.490 & 0.187 ± 0.157 & 0.180 ± 0.153 \\
				\cline{1-6}
				\multicolumn{2}{l}{\texttt{ShapleyValueSampling}} \\ 
				& Handcrafted & 0.825 ± 0.027 & 0.000 ± 0.000 & 0.010 ± 0.000 & 0.010 ± 0.000 \\
				& Trained & 0.999 ± 0.253 & 0.218 ± 0.403 & 0.045 ± 0.033 & 0.034 ± 0.024 \\
				\cline{1-6}
				\multicolumn{2}{l}{\texttt{LIME} (Linear)} \\ 
				& Handcrafted & 1.972 ± 0.719 & 0.000 ± 0.000 & 0.064 ± 0.001 & 0.064 ± 0.001 \\
				& Trained & 55.196 ± 26.218 & 0.247 ± 0.424 & 0.072 ± 0.008 & 0.063 ± 0.005 \\
				\cline{1-6}
				\multicolumn{2}{l}{\texttt{LIME} (Lasso)} \\ 
				& Handcrafted & 2.796 ± 1.055 & 0.004 ± 0.000 & 0.058 ± 0.001 & 0.059 ± 0.001 \\
				& Trained & 2.825 ± 0.877 & 0.230 ± 0.403 & 0.076 ± 0.016 & 0.068 ± 0.011 \\
				\hline
				\multicolumn{2}{l}{\cellcolor{headercolor} Model Performance\textsuperscript{\textit{j}}} &\cellcolor{headercolor} &\cellcolor{headercolor} & \cellcolor{headercolor} & \cellcolor{headercolor} \\
				\cellcolor{headercolor}
				& \cellcolor{headercolor} Handcrafted & \cellcolor{headercolor} 0.000 ± 0.000$^{\downarrow}$ & \cellcolor{headercolor} 0.000 ± 0.000$^{\downarrow}$ & \cellcolor{headercolor} 0.000 ± 0.000$^{\downarrow}$ & \cellcolor{headercolor} 0.000 ± 0.000$^{\downarrow}$ \\
				\cellcolor{headercolor}
				& \cellcolor{headercolor} Trained & \cellcolor{headercolor} 0.003 ± 0.003$^{\downarrow}$ & \cellcolor{headercolor} 1.100 ± 2.022$^{\downarrow}$ & \cellcolor{headercolor} 0.001 ± 0.003$^{\downarrow}$ & \cellcolor{headercolor} 0.002 ± 0.003$^{\downarrow}$ \\
				\cline{1-6}   
		\end{tabular}}\\
		\small \textsuperscript{\textit{a}}~{\texttt{WeightedFeaturesDataset}~(MSE$^{\downarrow}$)}\\
		\small \textsuperscript{\textit{b}}~{\texttt{ConflictingDataset}~(MSE$^{\downarrow}$)}\\
		\small \textsuperscript{\textit{c}}~{\texttt{InteractingFeatureDataset}~(MSE$^{\downarrow}$)}\\
		\small \textsuperscript{\textit{d}}~{\texttt{UncertaintyAwareDataset}~(Mask Error$^{\downarrow}$)}\\
		\small \textsuperscript{\textit{e}}~{The metric for Model Performance is MSE$^{\downarrow}$ except for \texttt{UncertaintyAwareDataset} using Accuracy$^{\uparrow}$.}\\
		\small \textsuperscript{\textit{f}}~{\texttt{ShatteredGradientsDataset}~(\texttt{SensitivityMax}$^{\downarrow}$)}\\
		\small \textsuperscript{\textit{g}}~{\texttt{PertinentNegativesDataset}~(MSE$^{\downarrow}$)}\\
		\small \textsuperscript{\textit{h}}~{\texttt{BooleanAndDataset}~(MSE$^{\downarrow}$)}\\
		\small \textsuperscript{\textit{i}}~{\texttt{BooleanOrDataset}~(MSE$^{\downarrow}$)}\\
		\small \textsuperscript{\textit{j}}~{The metric for Model Performance is MSE$^{\downarrow}$ except for \texttt{UncertaintyAwareDataset} using Accuracy$^{\uparrow}$.}\\
	\end{table*}
	
	\paragraph{Experiment setup} The FA methods we experimented with are (the \textsc{Captum}~\citep{kokhlikyan2020captum} versions of) \texttt{DeepLIFT}~\citep{shrikumar2019learning}, \texttt{InputXGradient}~\citep{shrikumar2017just}, \texttt{IntegratedGradients}~\citep{sundararajan2017axiomatic}, \texttt{LIME}~\citep{ribeiro2016why} (with linear regression without regularisation as the surrogate model, and with lasso regression), \texttt{KernelSHAP}~\citep{lundberg2017unified} and \texttt{ShapleyValueSampling}~\citep{castro2009polynomial}.
	We initialise the datasets introduced in Section ~\ref{sec:package} and split them into train, validate and test subsets (with 2600, 400, and 1000 data points respectively). We use the test subset for the FA evaluations. Each dataset has ten input features (except the \texttt{ConflictingDataset}, which has ten additional ``cancellation'' features) with 1000 data points for evaluation. The trained model was a \texttt{ReLU} MLP with three hidden layers, each 100 neurons wide. FA methods were evaluated using the default metric of each dataset. This experiment was repeated for five trials using different random model initialisations. All models were trained on a single retail GPU (RTX 4070Ti) and experiments were run on a single retail CPU (AMD Ryzen 9 7950X).
	
	\paragraph{High-level findings}
	Table~\ref{tab:tabular_results} summarises the results. We have given the mean and standard deviations for FA methods evaluated across the five trials and the model performance for the trained models (at the bottom of the table). All FA methods performed well on the simplest test case, the Weighted Continuous models, but struggled on models with gradient discontinuities, such as those for Shattered Gradients and Pertinent Negatives. In addition, in line with intuition, FA methods that rely on linear surrogate models such as \texttt{KernelSHAP} and \texttt{LIME} tend to perform worse on significantly non-linear models. Our results also indicate that FA methods generally perform worse on the trained models compared to the handcrafted models (with notable exceptions as discussed in the case studies below). This is expected, as imperfect, trained models may not be fully aligned with the ground truth. Thus, given the higher prediction error of the trained models (see Model Performance score) we would expect a higher attribution error. We also note the poor results for \texttt{InputXGradients} on certain models — this is likely due to gradients only being representative of the local model behaviour and not capturing the full effects of the input features. Following these general
	observations, some specific results may be of interest.
	
	\paragraph{Case study: Conflicting behaviour} Running the experiments on the Conflicting Feature models, we notice that gradients-based FA methods performed significantly worse on the handcrafted models in comparison with the trained models. We hypothesise this is due to the inherent limitations of gradient-based FA methods, which are exacerbated by our handcrafted model.
	
	Gradient-based FA methods are known to struggle with propagating importance signals when gradients are zero~\citep{shrikumar2019learning}. In our handcrafted Conflicting Feature model implementation (see Section~\ref{sec:package} and Appendix~\ref{sec:appendix_conflicting} for details), conflicting features push the gradients of the hidden layers to zero, resulting in zero attribution for both features involved in the conflict. This issue, zero-gradients leading to zero attribution scores, is also prevalent in any neural network with \texttt{ReLU} layers, including the trained models used. However, we observe that the effect is less pronounced in trained neural networks. We speculate that this is likely due to the conflict behaviour being distributed across multiple neurons, which would reduce the prevalence of truncated gradient signals during backpropagation.
	In contrast, perturbation-based methods do not seem to be affected by this problem. \texttt{ShapleyValueSampling}, in particular, is one of the best-performing methods for Conflicting Feature models. This phenomenon is likely attributable to its considerations of various combinations of inputs, which can reveal possible feature conflicts even if the network gradients are zero. However, merely relying on perturbations for computing attributions does not appear to be sufficient for accurately capturing the effects of conflicting features. Notably, FA methods using linear surrogate models also struggle with the Conflicting Feature models due to the non-linearity of the model output.

	\paragraph{Case study: Implementation discrepancy}
	The experiments run on the Uncertainty models highlight a discrepancy between the original design of \texttt{DeepLIFT} and the \textsc{Captum} implementation. The \texttt{DeepLIFT} paper recommends that, in the case of Softmax outputs, we may prefer to compute contributions to the logits rather than contributions to the Softmax outputs. \textit{If} we compute the contributions to the logits, then the paper also recommends applying a normalisation step. However, the \textsc{Captum} implementation applies the normalisation step whenever we compute contributions to the Softmax outputs and does not apply the normalisation step when computing contributions to the logits.
	
	We expect \texttt{DeepLift} to achieve a minimal Mask Error on the \texttt{UncertaintyAwareDataset} but the default \textsc{Captum} implementation gives a high score (Table~\ref{tab:deeplift}). By using the correct target layer for the attributions (logits) and applying the normalisation step, we see a perfect score. Our analysis demonstrates the utility of \textsc{XAI-Units} not only for comparing different FA methods, but also for surfacing and diagnosing issues with their implementation.

	\begin{table}[htb]    
		\caption{\texttt{DeepLIFT} on \texttt{UncertaintyAwareDataset}. Varying the target. $\downarrow / \uparrow $ indicates a low / high score is better.}
		\label{tab:deeplift}     
		\begin{center}
			\begin{tabular}{lcccc}
				\toprule
				& \multicolumn{3}{c}{Mask Error$^\downarrow$} \\
				&  &  & Logits with  \\
				& Outputs\textsuperscript{\textit{a}} &  Logits & Normalisation \\
				\midrule
				\multicolumn{1}{l}{Handcrafted} & 62.9 ± 80.5 & 0.991 ± 0.016 & 0.000 ± 0.000 \\
				\bottomrule
			\end{tabular}
			\small\textsuperscript{\textit{a}}~{\textsc{Captum} default implementation includes a normalisation bug.}
		\end{center}       
	\end{table}

	\subsection{Image dataset experiments}
	
	\begin{table*}[!htb]
		\caption{Image dataset results. $\downarrow / \uparrow $ indicates a low / high score is better.}
		\centering
		\begin{tabular}{lcccc}
			\toprule
			& \multicolumn{4}{c}{Mask Proportion Image$^\uparrow$} \\
			& \multicolumn{2}{c}{CNN} & \multicolumn{2}{c}{ViT} \\
			& Balanced Images & Imbalanced Images  & Balanced Images & Imbalanced Images \\
			\midrule
			\texttt{DeepLIFT} & 0.944 ± 0.048 & 0.522 ± 0.241 &0.708 ± 0.024     &     0.575 ± 0.040 \\
			\texttt{InputXGradient} & 0.955 ± 0.041 & 0.497 ± 0.209   &  0.708 ± 0.024     &     0.575 ± 0.040\\
			\texttt{IntegratedGradients} & 0.949 ± 0.046 & 0.531 ± 0.215  &  0.675 ± 0.019    &      0.491 ± 0.016 \\
			\hline
			\rowcolor{headercolor}
			Test Accuracy$^\uparrow$ & 0.862 ± 0.120 & 0.925 ± 0.112 & 0.818  ± 0.122 & 0.958  ± 0.089 \\
			\cline{1-5}
		\end{tabular}%
		\label{tab:image_results_CNN}
	\end{table*}
	
	Briefly, we also consider results on the image datasets, balanced and unbalanced, summarised by Table~\ref{tab:image_results_CNN}. The same regime was used for both datasets --- generating 3000 images with an 80/10/10 split where the test set is used for evaluating FA methods. A 1 million-parameter CNN and a Vision Transformer (ViT) with about 2 million parameters were used in the experiment, both randomly initialized across 5 seeds. These models were chosen to compare how different architectural designs impact performance and interpretability. CNNs rely on convolutional operations, which introduce inductive biases allowing them to efficiently detect local features across the image. In contrast, ViTs use attention mechanisms to capture global relationships between image patches but lack these inductive biases \citep{xu2021vitae}, making them less naturally translation invariant. As might be expected, for the imbalanced dataset, the model learns to rely on the background in order to achieve higher accuracy. As a consequence, we see that the FA methods assign a large proportion of the FA to the background rather than to the foreground shape. A CNN model trained with balanced backgrounds does not experience this problem so more than $90\%$ of the attributions are inside the shape mask. However, a ViT trained on balanced backgrounds shows significantly worse results, with approximately $70\%$ of the attributions falling inside the shape mask. We hypothesize that this might be caused by a lack of inductive bias for the locality in ViT, as the model can freely attend to arbitrary input patches instead of integrating information from progressively larger neighbourhoods as done by a CNN. The results allow us to compare how well different feature attribution methods discriminate between models focused on the true signal and those relying on spurious correlations with the background.

	\subsection{Text dataset experiments}
	
	\begin{table*}[!htb]
		\caption{Text dataset results. $\downarrow / \uparrow $ indicates a low/high score is better.}
		\centering
		\begin{tabular}{lccc}
			\toprule
			& Samples & \multicolumn{2}{c}{Mask Ratio$^\uparrow$} \\
			& & \texttt{TriggerLLM} & \texttt{TriggerLLM\textsubscript{Deterministic}} \\
			\midrule
			\texttt{FeatureAblation} & 1000 & 0.078 ± 0.102 & 1.000 ± 0.000\\
			\texttt{IntegratedGradients} & 1000 & -0.006 ± 0.138 & 0.040 ± 0.239\\
			\texttt{LIME} & 1000 & 0.018 ± 0.056 & 0.367 ± 0.113 \\
			\texttt{KernelSHAP} & 1000 & 0.013 ± 0.077 & 0.135 ± 0.168\\
			\texttt{ShapleyValueSampling} &  100 & 0.057 ± 0.063 & 0.669 ± 0.066\\        
			\bottomrule        
		\end{tabular}
		\label{tab:text_results}
	\end{table*}
	
	In the experiment with textual data, we compared five FA Methods, using our text dataset with two versions of the fine-tuned LLM: \texttt{TriggerLLM} and \texttt{TriggerLLM\textsubscript{Deterministic}}. The latter model is optimised to more reliably respond to the trigger token at the expense of a more substantial drop in performance on other tasks — see Appendix~\ref{sec:text_model} for more information. We focused on evaluating the FA methods applicable to LLMs using the official \textsc{Captum} wrappers. In all cases, the baseline token chosen was set to `` " or white-space token rather than the zero vector or zero token id, as we considered this to be a more natural and neutral baseline for textual data.
	
	\texttt{FeatureAblation} and \texttt{ShapleyValueSampling} were the two best-performing FA Methods for both LLMs. However, it is worth noting that \texttt{ShapleyValueSampling} takes an order of magnitude more time to run compared to other FA methods, hence only 100 samples were used for its experiment. The FA Methods based on linear surrogate models struggled due to the non-linear nature of the model/dataset but still produced reasonable attribution scores.
	
	\texttt{IntegratedGradients} was the worst-performing method. As \citet{sanyal2021discretizedintegratedgradientsexplaining} mention, the straight-line interpolation, used in \texttt{IntegratedGradients}, may not be appropriate given LLM inputs are discrete units and there are no intermediate states between two tokens, which can lead to inaccurate attributions.
	
	It is also noteworthy that the relative ranking of the FA methods evaluated on the two different LLMs was consistent.

	\section{Conclusion and discussion} \label{sec:conclusion}
	
	Within the XAI community, there is currently no consensus on the universally best approach for evaluating FA methods. While there are many existing benchmarks for this purpose, the benchmark we developed is unique in its focus on atomic ``test cases'' and comparison with ground-truth attribution scores. We achieve this by providing pairs of synthetic datasets and handcrafted neural network models. This creates a set of calibrated input interactions and model behaviours, against which we can evaluate FA methods using a battery of evaluation metrics. We do not claim that our benchmark offers a definitive ranking of FA methods. Rather, we argue that it is an effort towards addressing the \textit{disagreement problem}~\citep{krishna2022disagreement} and provides valuable insights into specific behaviours that can naturally complement results from other benchmarks and evaluation approaches. These insights can inform users about the strengths and weaknesses of FA methods in specific settings, improving transparency in their application. Furthermore, isolating scenarios where a particular FA method may falter can verify whether the design specifications of the method are fulfilled, thereby promoting accountability for the developer. This is exemplified by our case study in Section~\ref{sec:tabular_experiments}, which highlights the discrepancy in the implementation of \textsc{DeepLIFT} in \textsc{Captum}.
	
	Our benchmark is accessible in the form of the \textsc{XAI-Units} Python package, which is fully open-source and extensible to custom feature attribution methods or evaluation metrics. Moreover, \textsc{XAI-Units} has been designed such that the evaluation procedure is streamlined and researchers can effortlessly run their experiments on multiple FA methods.
	
	A potential limitation of our work is that the performance of FA methods on handcrafted neural networks is not guaranteed to represent their performance in the real world~\citep{rahnama2023blame}. Nevertheless, we believe in the merits of our approach. First, using synthetic models is the only way to guarantee model alignment with the data distribution and expected behaviour. Second, our models are directly modelling real-world scenarios, e.g., interacting or conflicting features. Finally, we also provide trained models for comparison, which enables us to see how real networks behave under controlled conditions while loosening the alignment guarantees.
	
	While the experiments conducted in this report showcase that our benchmark can be used to evaluate the correctness of FA methods, we note that other properties of FA methods, such as the \textit{Co-12} properties introduced by \citet{nauta2023}, need to be evaluated to ensure holistic assessment of FA methods. Although our benchmark is designed to be compatible with custom FA methods evaluation metrics (thus supports multi-faceted evaluations), ultimately it is limited to a technical evaluation of FA methods. 
	
	Human interpretability is an increasingly important property of XAI methods~\citep{kim2024} and many additional factors need to be considered for human-centric evaluations, such as explanation complexity (affecting how understandable they are to humans) and the role of explanations in effective human-AI interaction. Addressing these aspects may require user studies and human-AI performance evaluations, which are out of the scope of our current benchmark. As potential future work, developing a user-friendly graphical interface could enhance the accessibility and usability of our benchmark for a broader range of researchers and practitioners.
	
	Overall, to the best of our knowledge, our work is the first within the research community that provides an end-to-end pipeline to benchmark FA methods via a diverse set of synthetic datasets and handcrafted models. With ease of use and transparency in modelling, we hope that our work will aid researchers and practitioners alike in gaining a better understanding of the strengths and limitations of various FA methods. 
	
	\subsubsection*{Acknowledgements}
	We thank Theo Reynolds for his involvement and contributions throughout the development of this project.
	
	This research was partially supported by ERC under the EU's Horizon 2020 research and innovation programme (grant agreement no. 101020934, ADIX), by J.P. Morgan and the Royal Academy of Engineering under the Research Chairs and Senior Research Fellowships scheme (grant agreement no. RCSRF2021\textbackslash11\textbackslash45) and by UKRI through the CDT in AI for Healthcare \url{https://ai4health.io/} (grant agreement no. EP/S023283/1).
	
	\bibliography{references}

\begin{thebibliography}{}

\bibitem[Agarwal et~al., 2022]{agarwal2024openxai}
Agarwal, C., Krishna, S., Saxena, E., Pawelczyk, M., Johnson, N., Puri, I.,
  Zitnik, M., and Lakkaraju, H. (2022).
\newblock {OpenXAI}: Towards a transparent evaluation of model explanations.
\newblock In Koyejo, S., Mohamed, S., Agarwal, A., Belgrave, D., Cho, K., and
  Oh, A., editors, {\em Advances in Neural Information Processing Systems 35:
  Annual Conference on Neural Information Processing Systems 2022, NeurIPS
  2022, New Orleans, LA, USA, November 28 - December 9, 2022}.

\bibitem[Agarwal et~al., 2023]{agarwal2023evaluating}
Agarwal, C., Queen, O., Lakkaraju, H., and Zitnik, M. (2023).
\newblock Evaluating explainability for graph neural networks.
\newblock {\em Scientific Data}, 10(144).

\bibitem[Alvarez{-}Melis and Jaakkola, 2018]{alvarezmelis2018robustness}
Alvarez{-}Melis, D. and Jaakkola, T.~S. (2018).
\newblock On the robustness of interpretability methods.

\bibitem[Ancona et~al., 2018]{ancona2018better}
Ancona, M., Ceolini, E., {\"{O}}ztireli, C., and Gross, M. (2018).
\newblock Towards better understanding of gradient-based attribution methods
  for deep neural networks.
\newblock In {\em 6th International Conference on Learning Representations,
  {ICLR} 2018, Vancouver, BC, Canada, April 30 - May 3, 2018, Conference Track
  Proceedings}. OpenReview.net.

\bibitem[Arras et~al., 2022]{arras2022clevr}
Arras, L., Osman, A., and Samek, W. (2022).
\newblock {CLEVR-XAI}: A benchmark dataset for the ground truth evaluation of
  neural network explanations.
\newblock {\em Information Fusion}, 81:14--40.

\bibitem[Bach et~al., 2015]{bach2015pixel}
Bach, S., Binder, A., Montavon, G., Klauschen, F., M{\"u}ller, K.-R., and
  Samek, W. (2015).
\newblock On pixel-wise explanations for non-linear classifier decisions by
  layer-wise relevance propagation.
\newblock {\em PloS one}, 10(7):e0130140.

\bibitem[Bilodeau et~al., 2024]{Bilodeau_2024}
Bilodeau, B., Jaques, N., Koh, P.~W., and Kim, B. (2024).
\newblock Impossibility theorems for feature attribution.
\newblock {\em Proceedings of the National Academy of Sciences}, 121(2).

\bibitem[Breiman, 2001]{breiman2001statistical}
Breiman, L. (2001).
\newblock Statistical modeling: The two cultures (with comments and a rejoinder
  by the author).
\newblock {\em Statistical science}, 16(3):199--231.

\bibitem[Castro et~al., 2009]{castro2009polynomial}
Castro, J., G{\'o}mez, D., and Tejada, J. (2009).
\newblock Polynomial calculation of the shapley value based on sampling.
\newblock {\em Computers \& Operations Research}, 36(5):1726--1730.

\bibitem[Cimpoi et~al., 2014]{cimpoi14describing2014}
Cimpoi, M., Maji, S., Kokkinos, I., Mohamed, S., and Vedaldi, A. (2014).
\newblock Describing textures in the wild.
\newblock In {\em 2014 {IEEE} Conference on Computer Vision and Pattern
  Recognition, {CVPR} 2014, Columbus, OH, USA, June 23-28, 2014}, pages
  3606--3613. {IEEE} Computer Society.

\bibitem[Cobbe et~al., 2021]{cobbe2021gsm8k}
Cobbe, K., Kosaraju, V., Bavarian, M., Chen, M., Jun, H., Kaiser, L., Plappert,
  M., Tworek, J., Hilton, J., Nakano, R., Hesse, C., and Schulman, J. (2021).
\newblock Training verifiers to solve math word problems.
\newblock {\em CoRR}, abs/2110.14168.

\bibitem[Cui et~al., 2022]{cui-etal-2022-expmrc}
Cui, Y., Liu, T., Che, W., Chen, Z., and Wang, S. (2022).
\newblock Expmrc: Explainability evaluation for machine reading comprehension.
\newblock {\em Heliyon}, 8:e09290.

\bibitem[Dabkowski and Gal, 2017]{dabkowski2017real}
Dabkowski, P. and Gal, Y. (2017).
\newblock Real time image saliency for black box classifiers.
\newblock In Guyon, I., von Luxburg, U., Bengio, S., Wallach, H.~M., Fergus,
  R., Vishwanathan, S. V.~N., and Garnett, R., editors, {\em Advances in Neural
  Information Processing Systems 30: Annual Conference on Neural Information
  Processing Systems 2017, December 4-9, 2017, Long Beach, CA, {USA}}, pages
  6967--6976.

\bibitem[Dejl et~al., 2025]{dejl2025conflicts}
Dejl, A., Zhang, D., Ayoobi, H., Williams, M., and Toni, F. (2025).
\newblock Hidden conflicts in neural networks and their implications for
  explainability.
\newblock In {\em FAccT '25: The 2025 ACM Conference on Fairness,
  Accountability, and Transparency Proceedings}, New York, NY, USA. Association
  for Computing Machinery.
\newblock To appear.

\bibitem[Ding et~al., 2023]{ding2023enhancing}
Ding, N., Chen, Y., Xu, B., Qin, Y., Hu, S., Liu, Z., Sun, M., and Zhou, B.
  (2023).
\newblock Enhancing chat language models by scaling high-quality instructional
  conversations.
\newblock In Bouamor, H., Pino, J., and Bali, K., editors, {\em Proceedings of
  the 2023 Conference on Empirical Methods in Natural Language Processing},
  pages 3029--3051, Singapore. Association for Computational Linguistics.

\bibitem[Fresz et~al., 2024]{Fresz2024}
Fresz, B., L\"{o}rcher, L., and Huber, M. (2024).
\newblock Classification metrics for image explanations: Towards building
  reliable xai-evaluations.
\newblock In {\em Proceedings of the 2024 ACM Conference on Fairness,
  Accountability, and Transparency}, FAccT '24, page 1–19, New York, NY, USA.
  Association for Computing Machinery.

\bibitem[Grattafiori et~al., 2024]{grattafiori2024llama3herdmodels}
Grattafiori, A., Dubey, A., Jauhri, A., Pandey, A., Kadian, A., Al-Dahle, A.,
  Letman, A., Mathur, A., Schelten, A., Vaughan, A., Yang, A., Fan, A., et~al.
  (2024).
\newblock The {L}lama 3 herd of models.

\bibitem[Guidotti, 2021]{guidotti2021evaluating}
Guidotti, R. (2021).
\newblock Evaluating local explanation methods on ground truth.
\newblock {\em Artificial Intelligence}, 291:103428.

\bibitem[Hedstr{\"{o}}m et~al., 2023]{hedstrom2023quantus}
Hedstr{\"{o}}m, A., Weber, L., Krakowczyk, D., Bareeva, D., Motzkus, F., Samek,
  W., Lapuschkin, S., and H{\"{o}}hne, M.~M. (2023).
\newblock Quantus: An explainable {AI} toolkit for responsible evaluation of
  neural network explanations and beyond.

\bibitem[Hossein and Rahnama, 2024]{rahnama2023blame}
Hossein, A. and Rahnama, A. (2024).
\newblock The blame problem in evaluating local explanations and how to tackle
  it.
\newblock In {\em Artificial Intelligence. ECAI 2023 International Workshops},
  pages 66--86, Cham. Springer Nature Switzerland.

\bibitem[Huang et~al., 2023]{huang2023safari}
Huang, W., Zhao, X., Jin, G., and Huang, X. (2023).
\newblock {SAFARI}: Versatile and efficient evaluations for robustness of
  interpretability.
\newblock In {\em {IEEE/CVF} International Conference on Computer Vision,
  {ICCV} 2023, Paris, France, October 1-6, 2023}, pages 1988--1998. {IEEE}.

\bibitem[Kim et~al., 2024]{kim2024}
Kim, J., Maathuis, H., and Sent, D. (2024).
\newblock Human-centered evaluation of explainable ai applications: A
  systematic review.
\newblock {\em Frontiers in Artificial Intelligence}, Volume 7 - 2024.

\bibitem[Kokhlikyan et~al., 2020]{kokhlikyan2020captum}
Kokhlikyan, N., Miglani, V., Martin, M., Wang, E., Alsallakh, B., Reynolds, J.,
  Melnikov, A., Kliushkina, N., Araya, C., Yan, S., and Reblitz{-}Richardson,
  O. (2020).
\newblock Captum: {A} unified and generic model interpretability library for
  {PyTorch}.

\bibitem[Krishna et~al., 2024]{krishna2022disagreement}
Krishna, S., Han, T., Gu, A., Wu, S., Jabbari, S., and Lakkaraju, H. (2024).
\newblock The disagreement problem in explainable machine learning: {A}
  practitioner's perspective.
\newblock {\em Trans. Mach. Learn. Res.}, 2024.

\bibitem[Le et~al., 2023]{le2023benchmarking}
Le, P.~Q., Nauta, M., Nguyen, V.~B., Pathak, S., Schl{\"{o}}tterer, J., and
  Seifert, C. (2023).
\newblock Benchmarking explainable {AI} - {A} survey on available toolkits and
  open challenges.
\newblock In {\em Proceedings of the Thirty-Second International Joint
  Conference on Artificial Intelligence, {IJCAI} 2023, 19th-25th August 2023,
  Macao, SAR, China}, pages 6665--6673. ijcai.org.

\bibitem[Li et~al., 2023]{li2023mathcal}
Li, X., Du, M., Chen, J., Chai, Y., Lakkaraju, H., and Xiong, H. (2023).
\newblock $\mathcal{M}^4$: A unified {XAI} benchmark for faithfulness
  evaluation of feature attribution methods across metrics, modalities and
  models.
\newblock {\em Advances in Neural Information Processing Systems},
  36:1630--1643.

\bibitem[Lin et~al., 2021]{lin2020see}
Lin, Y., Lee, W., and Celik, Z.~B. (2021).
\newblock What do you see?: Evaluation of explainable artificial intelligence
  {(XAI)} interpretability through neural backdoors.
\newblock In Zhu, F., Ooi, B.~C., and Miao, C., editors, {\em {KDD} '21: The
  27th {ACM} {SIGKDD} Conference on Knowledge Discovery and Data Mining,
  Virtual Event, Singapore, August 14-18, 2021}, pages 1027--1035. {ACM}.

\bibitem[Liu et~al., 2021]{liu2021synthetic}
Liu, Y., Khandagale, S., White, C., and Neiswanger, W. (2021).
\newblock Synthetic benchmarks for scientific research in explainable machine
  learning.
\newblock In Vanschoren, J. and Yeung, S., editors, {\em Proceedings of the
  Neural Information Processing Systems Track on Datasets and Benchmarks 1,
  NeurIPS Datasets and Benchmarks 2021, December 2021, virtual}.

\bibitem[Lundberg and Lee, 2017]{lundberg2017unified}
Lundberg, S.~M. and Lee, S. (2017).
\newblock A unified approach to interpreting model predictions.
\newblock In Guyon, I., von Luxburg, U., Bengio, S., Wallach, H.~M., Fergus,
  R., Vishwanathan, S. V.~N., and Garnett, R., editors, {\em Advances in Neural
  Information Processing Systems 30: Annual Conference on Neural Information
  Processing Systems 2017, December 4-9, 2017, Long Beach, CA, {USA}}, pages
  4765--4774.

\bibitem[Mamalakis et~al., 2022]{mamalakis2022neural}
Mamalakis, A., Ebert-Uphoff, I., and Barnes, E.~A. (2022).
\newblock Neural network attribution methods for problems in geoscience: A
  novel synthetic benchmark dataset.
\newblock {\em Environmental Data Science}, 1:e8.

\bibitem[Nauta et~al., 2023]{nauta2023}
Nauta, M., Trienes, J., Pathak, S., Nguyen, E., Peters, M., Schmitt, Y.,
  Schlötterer, J., van Keulen, M., and Seifert, C. (2023).
\newblock From anecdotal evidence to quantitative evaluation methods: A
  systematic review on evaluating explainable ai.
\newblock {\em ACM Computing Surveys}, 55(13s):1–42.

\bibitem[Ramaswamy et~al., 2020]{ramaswamy2020ablation}
Ramaswamy, H.~G. et~al. (2020).
\newblock {Ablation-CAM}: Visual explanations for deep convolutional network
  via gradient-free localization.
\newblock In {\em proceedings of the IEEE/CVF winter conference on applications
  of computer vision}, pages 983--991.

\bibitem[Ribeiro et~al., 2016]{ribeiro2016why}
Ribeiro, M.~T., Singh, S., and Guestrin, C. (2016).
\newblock {"Why} should {I} trust you?": Explaining the predictions of any
  classifier.
\newblock In Krishnapuram, B., Shah, M., Smola, A.~J., Aggarwal, C.~C., Shen,
  D., and Rastogi, R., editors, {\em Proceedings of the 22nd {ACM} {SIGKDD}
  International Conference on Knowledge Discovery and Data Mining, San
  Francisco, CA, USA, August 13-17, 2016}, pages 1135--1144. {ACM}.

\bibitem[Roy et~al., 2022]{roy2022}
Roy, S., Laberge, G., Roy, B., Khomh, F., Nikanjam, A., and Mondal, S. (2022).
\newblock Why don’t {XAI} techniques agree? {Characterizing} the
  disagreements between post-hoc explanations of defect predictions.
\newblock In {\em 2022 IEEE International Conference on Software Maintenance
  and Evolution (ICSME)}, pages 444--448.

\bibitem[Saha et~al., 2020]{saha2019hiddentriggerbackdoorattacks}
Saha, A., Subramanya, A., and Pirsiavash, H. (2020).
\newblock Hidden trigger backdoor attacks.
\newblock In {\em The Thirty-Fourth {AAAI} Conference on Artificial
  Intelligence, {AAAI} 2020, The Thirty-Second Innovative Applications of
  Artificial Intelligence Conference, {IAAI} 2020, The Tenth {AAAI} Symposium
  on Educational Advances in Artificial Intelligence, {EAAI} 2020, New York,
  NY, USA, February 7-12, 2020}, pages 11957--11965. {AAAI} Press.

\bibitem[Sanyal and Ren,
  2021]{sanyal2021discretizedintegratedgradientsexplaining}
Sanyal, S. and Ren, X. (2021).
\newblock Discretized integrated gradients for explaining language models.
\newblock In Moens, M.-F., Huang, X., Specia, L., and Yih, S. W.-t., editors,
  {\em Proceedings of the 2021 Conference on Empirical Methods in Natural
  Language Processing}, pages 10285--10299, Online and Punta Cana, Dominican
  Republic. Association for Computational Linguistics.

\bibitem[Shrikumar et~al., 2017]{shrikumar2019learning}
Shrikumar, A., Greenside, P., and Kundaje, A. (2017).
\newblock Learning important features through propagating activation
  differences.
\newblock In Precup, D. and Teh, Y.~W., editors, {\em Proceedings of the 34th
  International Conference on Machine Learning, {ICML} 2017, Sydney, NSW,
  Australia, 6-11 August 2017}, volume~70 of {\em Proceedings of Machine
  Learning Research}, pages 3145--3153. {PMLR}.

\bibitem[Shrikumar et~al., 2016]{shrikumar2017just}
Shrikumar, A., Greenside, P., Shcherbina, A., and Kundaje, A. (2016).
\newblock Not just a black box: Learning important features through propagating
  activation differences.

\bibitem[Speith, 2022]{speith2022}
Speith, T. (2022).
\newblock A review of taxonomies of explainable artificial intelligence {(XAI)}
  methods.
\newblock In {\em Proceedings of the 2022 ACM Conference on Fairness,
  Accountability, and Transparency}, FAccT '22, page 2239–2250, New York, NY,
  USA. Association for Computing Machinery.

\bibitem[Sundararajan et~al., 2017]{sundararajan2017axiomatic}
Sundararajan, M., Taly, A., and Yan, Q. (2017).
\newblock Axiomatic attribution for deep networks.
\newblock In Precup, D. and Teh, Y.~W., editors, {\em Proceedings of the 34th
  International Conference on Machine Learning, {ICML} 2017, Sydney, NSW,
  Australia, 6-11 August 2017}, volume~70 of {\em Proceedings of Machine
  Learning Research}, pages 3319--3328. {PMLR}.

\bibitem[{Wikimedia Commons}, 2024]{wiki2024}
{Wikimedia Commons} (2024).
\newblock {Category:Dinosaurs} with transparent background.
\newblock Online; accessed 2-June-2024.

\bibitem[Xu et~al., 2021]{xu2021vitae}
Xu, Y., Zhang, Q., Zhang, J., and Tao, D. (2021).
\newblock {ViTAE}: Vision transformer advanced by exploring intrinsic inductive
  bias.
\newblock In Ranzato, M., Beygelzimer, A., Dauphin, Y.~N., Liang, P., and
  Vaughan, J.~W., editors, {\em Advances in Neural Information Processing
  Systems 34: Annual Conference on Neural Information Processing Systems 2021,
  NeurIPS 2021, December 6-14, 2021, virtual}, pages 28522--28535.

\bibitem[Yan et~al., 2024]{yan2024backdooringinstructiontunedlargelanguage}
Yan, J., Yadav, V., Li, S., Chen, L., Tang, Z., Wang, H., Srinivasan, V., Ren,
  X., and Jin, H. (2024).
\newblock Backdooring instruction-tuned large language models with virtual
  prompt injection.
\newblock In Duh, K., Gomez, H., and Bethard, S., editors, {\em Proceedings of
  the 2024 Conference of the North American Chapter of the Association for
  Computational Linguistics: Human Language Technologies (Volume 1: Long
  Papers)}, pages 6065--6086, Mexico City, Mexico. Association for
  Computational Linguistics.

\bibitem[Yang and Kim, 2019]{yang2019benchmarking}
Yang, M. and Kim, B. (2019).
\newblock Benchmarking attribution methods with relative feature importance.
\newblock {\em CoRR}, abs/1907.09701.

\bibitem[Yeh et~al., 2019]{yeh2019infidelity}
Yeh, C.-K., Hsieh, C.-Y., Suggala, A.~S., Inouye, D.~I., and Ravikumar, P.
  (2019).
\newblock On the (in)fidelity and sensitivity of explanations.
\newblock In {\em Proceedings of the 33rd International Conference on Neural
  Information Processing Systems}, Red Hook, NY, USA. Curran Associates Inc.

\bibitem[Zhang et~al., 2023a]{zhang2023xai}
Zhang, Y., Gu, S., Song, J., Pan, B., and Zhao, L. (2023a).
\newblock {XAI} benchmark for visual explanation.

\bibitem[Zhang et~al., 2023b]{zhang2024attributionlab}
Zhang, Y., Li, Y., Brown, H., Rezaei, M., Bischl, B., Torr, P. H.~S., Khakzar,
  A., and Kawaguchi, K. (2023b).
\newblock {AttributionLab}: Faithfulness of feature attribution under
  controllable environments.

\bibitem[Zhou et~al., 2021]{zhou2021evaluating}
Zhou, J., Gandomi, A.~H., Chen, F., and Holzinger, A. (2021).
\newblock Evaluating the quality of machine learning explanations: A survey on
  methods and metrics.
\newblock {\em Electronics}, 10(5):593.

\bibitem[Zhou et~al., 2022]{zhou2022feature}
Zhou, Y., Booth, S., Ribeiro, M.~T., and Shah, J. (2022).
\newblock Do feature attribution methods correctly attribute features?
\newblock In {\em Thirty-Sixth {AAAI} Conference on Artificial Intelligence,
  {AAAI} 2022, Thirty-Fourth Conference on Innovative Applications of
  Artificial Intelligence, {IAAI} 2022, The Twelveth Symposium on Educational
  Advances in Artificial Intelligence, {EAAI} 2022 Virtual Event, February 22 -
  March 1, 2022}, pages 9623--9633. {AAAI} Press.

\end{thebibliography}
	
	\clearpage
	
	\appendix
	\onecolumn
	
	\section{Dataset and model definitions} \label{sec:nn_diagrams_appendix}
	
\begin{figure}[htb]
		\centering
		\includegraphics[width=0.3\linewidth]{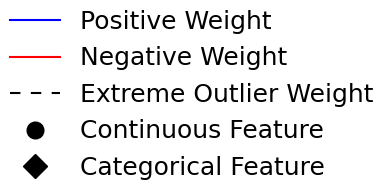} 
		\caption{Legend for the model diagrams in Appendix~\ref{sec:nn_diagrams_appendix}. The diagrams illustrate one instance of each model but note that the number of input features can be adjusted.}
		\label{fig:legend}
	\end{figure}

	This section details the formulae and diagram of the models for each of the dataset-model pairs that \textsc{XAI-Units} provide, as well as the details on the used ground-truth attributions. Figure~\ref{fig:legend} displays the legend for interpreting the subsequent diagrams.
	
	\subsection{Baselines}
	We first briefly introduce the used attribution baselines. Most FA methods calculate attribution scores relative to a baseline input~\citep{sundararajan2017axiomatic,shrikumar2017just,shrikumar2019learning}. The consideration of a baseline has been argued to make feature attribution more flexible and enable them to consider the full effects of the input features instead of merely focusing on local variations of the model function over small regions. Apart from their usage in several gradient-based methods, baselines are also relevant for the theoretically justified SHAP explanations ~\citep{lundberg2017unified}. Thus, we see it fitting and intuitive to calculate the ground-truth attributions with respect to a baseline (see the sub-sections below for precise derivations of the ground-truths for the individual models). Unless otherwise specified, we use the baseline of zero for all our attributions due to it being typically considered a neutral choice.

	\subsection{Weighted Continuous}
	
	\begin{figure}[!htb]
		\begin{minipage}[h]{0.5\textwidth}
			% \vspace{-\baselineskip}
			$$y = \mathbf{W}x = \sum^{n}_{i=1} w_ix_i$$
		\end{minipage}%
		\begin{minipage}[h]{0.38\textwidth}
			\centering
			\includegraphics[width=\linewidth]{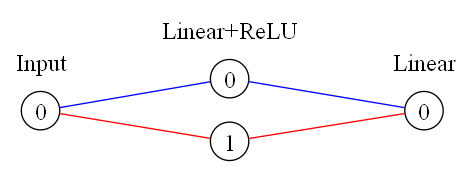} 
			% \captionof{figure}{ Continuous model diagram} 
			
		\end{minipage}
		\caption{\textbf{Weighted Continuous} formula and model diagram. For a feature vector $x \in \mathbb{R}^{n}$  and  a given weight matrix $\mathbf{W} \in \mathbb{R}^{1 \times n}$, where $n\in \mathbb{N}$ is the number of features, the output $y$ of the model.}
		\label{fig:continuous}
	\end{figure}
	
	The default evaluation metric is MSE, measuring the difference from the ground truth attributions. The ground truth feature attribution for continuous feature $x_i$ can be defined by ablating to $x_{ref} = 0$:
	
	\begin{equation*}
		\begin{aligned}
			FA_{x_i}(\mathbf{x}) &= M(\mathbf{x}) - M(\mathbf{x}_{-i}) \\
			\text{where } \mathbf{x} &= (x_1, \dots, x_n) \\
			\text{and } \mathbf{x}_{-i} &= (x_1, \dots, x_{i-1}, 0, x_{i+1}, \dots, x_n)
		\end{aligned}
	\end{equation*}
	
	\subsection{Conflicting Features}
	\label{sec:appendix_conflicting}
	\begin{figure}[!htb]
		\noindent
		\begin{minipage}[h]{0.5\textwidth}
			\vspace{-0.5\baselineskip}
			\begin{align*}
				y &= \sum^{n}_{i=1} z_i \\
				z_{i} &=  
				\begin{cases}
					w_i x_i & \text{if } c_i = 0\\
					0 & \text{if } c_i = 1 
				\end{cases}
			\end{align*}
		\end{minipage}%
		\begin{minipage}[h]{0.38\textwidth}
			\centering
			\includegraphics[width=\linewidth]{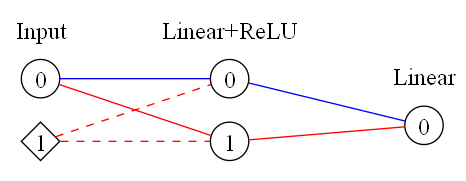} 
			% \captionof{figure}{Conflicting Model diagram} 
			% \label{subim1}
		\end{minipage}
		\caption{\textbf{Conflicting Features} formula and model diagram. For a continuous feature $x_i$ and a (categorical) cancellation feature $c_i$, together $(x_i, c_i)$ contribute to output $y$.}
		\label{fig:conflicting}
	\end{figure}
	
	The ground truth feature attribution is defined by ablating to a baseline reference $(x_{ref}, c_{ref}) = (0,0)$.
	
	\begin{equation*}
		\begin{aligned}
			FA_{c_i}(\mathbf{x},\mathbf{c}) &= M(\mathbf{x}, \mathbf{c}) - M(\mathbf{x}, \mathbf{c}_{-i}) \\
			FA_{x_i}(\mathbf{x},\mathbf{c}) &= M(\mathbf{x}, \mathbf{c}_{-i}) - M(\mathbf{x}_{-i},\mathbf{c}_{-i})
		\end{aligned}
	\end{equation*}
	
	\subsection{Pertinent Negatives}
	
	\begin{figure}[!htb]
		\noindent
		\begin{minipage}[h]{0.5\textwidth}
			\vspace{-\baselineskip}
			\begin{align*}
				y &= \sum^{n}_{i=1} z_i \\
				z_{i} &=
				\begin{cases}
					w_i (x_i + m(1-x_i))   & \textnormal{if } i \in P_i\\
					w_ix_i & \textnormal{otherwise} \\
					% w_ix_i & \text{if $i \notin \{$ pertinent negative features $\}$}\\
					% w_i   & \text{if $i \in \{$ pertinent negative features $\} \cup \ \{ x_i = 1 \}$  }\\
					% w_i*m  & \text{if $i \in \{$ pertinent negative features $\} \cup \ \{ x_i = 0 \}$  }\\
				\end{cases}
			\end{align*}
		\end{minipage}%
		\begin{minipage}[h]{0.5\textwidth}
			\centering
			\includegraphics[width=\linewidth]{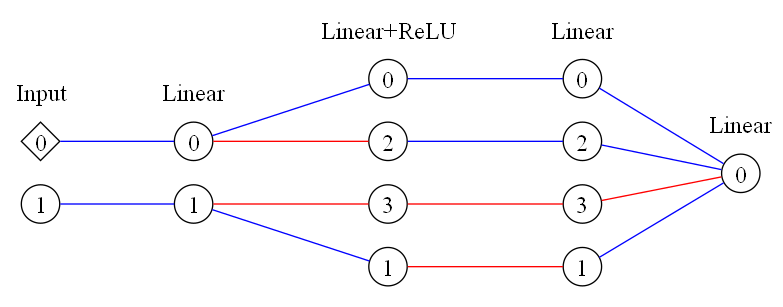}
			% \captionof{figure}{Pertinent negative diagram} 
			
		\end{minipage}
		\caption{\textbf{Pertinent Negative} formula and model diagram. $P_i$ denotes the set of indices of all pertinent negative features. For simplicity, we assume that pertinent negative features are categorical with values $0$ or $1$. When the pertinent negative feature $x_i$ takes a value of $0$, the output value is modified by a multiplier $m\in \mathbb{R}$.}
		\label{fig:PN_model}
	\end{figure}
	
	The ground truth feature attribution is defined by ablating to a baseline reference $x_{ref} = 0$.
	
	\begin{equation*}
		\begin{aligned}
			FA_{x_i}(\mathbf{x}) &= M(\mathbf{x}) - M(\mathbf{x}_{-i})
		\end{aligned}
	\end{equation*}
	
	\subsection{Shattered Gradients}
	
	\begin{figure}[!htb]
		\noindent
		\begin{minipage}[h]{0.5\textwidth}
			\begin{align*}
				y &= \sum^{n}_{i=1} \operatorname{ReLU}(z_i) \\ 
				z_i &= w_i x_i\\
			\end{align*}
		\end{minipage}%
		\begin{minipage}[h]{0.25\textwidth}
			\centering
			\includegraphics[width=\linewidth]{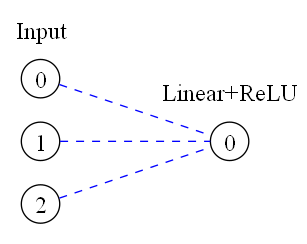}
			% \captionof{figure}{Shatterred Gradient diagram} 
			% \label{subim1}
		\end{minipage}
		\caption{\textbf{Shattered Gradients} formula and model diagram.}
		\label{Shaterred}
	\end{figure}
	
	Ground truth feature attributions are not available for the Shattered Gradients model. We use \texttt{SensitivityMax} as the default evaluation metric.
	
	\subsection{Categorical Feature Interaction}
	
	\begin{figure}[!htb]
		\noindent
		\begin{minipage}[c]{0.5\textwidth}
			\vspace{-\baselineskip}
			\begin{gather*}
				y = \sum^{n}_{i=1} z_i\\
				z_{i} =
				\begin{cases}
					w_{i}x_i & \text{if $x_i$ is non-interacting}\\
					% w^{(1)}_{i}x_i  & \text{if $x_i$ is interacting with categorical $c_i$ and $c_i = 0$}\\
					% w^{(2)}_{i}x_i  & \text{if $x_i$ is interacting with categorical $c_i$ and $c_i = 1$ }\\
					w_ic_i + x_i(w^{(1)}_{i}(1-c_i) + w^{(2)}_{i}c_i) & \text{if $x_i$ interacts with $c_i$}\\
				\end{cases}
			\end{gather*}
		\end{minipage}%
		\begin{minipage}[c]{0.3\textwidth}
			\centering
			\includegraphics[width=\linewidth]{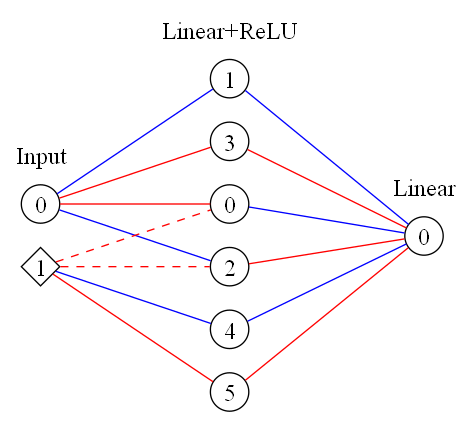}
			% \captionof{figure}{Interaction Model diagram} 
			% \label{subim1}
		\end{minipage}
		\caption{\textbf{Categorical Feature Interaction} formula and model diagram. The user can define some features to be non-interacting and other features to have an interaction. For an interacting pair $(x_i,c_i)$ then $c_i$ is categorical with value $0$ or $1$. If $c_i$ is $0$, the weight applied to the continuous feature $x_i$ is $w^{(1)}_i$. If $c_i$ is $1$, the weight applied to $x_i$ becomes $w^{(2)}_i$.}
		\label{fig:interacting}
	\end{figure}
	
	The ground truth feature attribution is defined by ablating to a baseline reference $(x_{ref}, c_{ref}) = (0,0)$.
	
	\begin{equation*}
		\begin{aligned}
			FA_{x_i}(\mathbf{x},\mathbf{c}) &= M(\mathbf{x}, \mathbf{c}) - M(\mathbf{x}_{-i},\mathbf{c}) \\
			FA_{c_i}(\mathbf{x},\mathbf{c}) &= M(\mathbf{x}_{-i}, \mathbf{c}) - M(\mathbf{x}_{-i}, \mathbf{c}_{-i})
		\end{aligned}
	\end{equation*}
	
	\subsection{Uncertainty Model}
	\label{sec:appendix_uncertainty}
	
	\begin{figure}[!htb]
		\noindent
		\begin{minipage}[h]{0.5\textwidth}
			\vspace{-\baselineskip}
			\begin{align*}
				y_i &= \operatorname{softmax}(z_i) \\
				z_i &= w_i x_i + \sum_{k} w_k x_k \\
				i &\in \{\text{index of \texttt{standard} features}\} \\
				k &\in \{\text{index of \texttt{common} features}\} 
			\end{align*}
		\end{minipage}%
		\begin{minipage}[h]{0.26\textwidth}
			\centering
			\includegraphics[width=\linewidth]{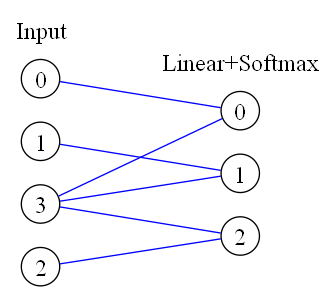}
			% \captionof{figure}{Uncertainty Model diagram} 
			% \label{subim1}
		\end{minipage}
		\caption{\textbf{Uncertainty Model} formula and model diagram.}
		\label{fig:uncertainty}
	\end{figure}
	
	The Uncertainty dataset is intended for classification problems rather than regression problems. For this dataset (like the image dataset) we provide the ground truth as a mask rather than exact feature attributions. The mask is defined as:
	
	\begin{displaymath}
		\begin{aligned}
			FA_{x_i} =
			\begin{cases}
				1 & \text{if $x_i$ is a \texttt{standard} feature}\\
				0    & \text{if $x_i$ is a \texttt{common} feature}\\
			\end{cases}
		\end{aligned}
	\end{displaymath}

	When scoring an FA method on the Uncertainty dataset, the default metric is not MSE but the Mask Error. This is calculated as the mean squared attribution assigned to the \texttt{common} features, so it gives a measure of how much attribution falls outside the mask.
	
	\subsection{Boolean Formulae}

	\begin{figure}[!htb]
		\centering
		\begin{subfigure}{0.4\textwidth}
			\centering
			\vspace*{\fill} % Add vertical space at the top
			\[
			\textnormal{NOT}(p) = -p
			\]
			\vspace*{\fill} % Add vertical space at the bottom
			\caption{Boolean NOT model}
			\label{fig:not}
		\end{subfigure}%
		\begin{subfigure}{0.6\textwidth}
			\centering
			\includegraphics[width=\linewidth]{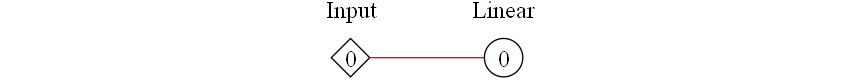}
			\vspace*{\fill}
		\end{subfigure}
		\vspace{3mm}
		
		\begin{subfigure}{0.4\textwidth}
			\centering
			\vspace*{\fill} % Add vertical space at the top
			\[
			\textnormal{OR}(p, q) = \frac{p+q + |p-q|}{2}
			\]
			\vspace*{\fill} % Add vertical space at the bottom
			\caption{Boolean OR model}
			\label{fig:or}
		\end{subfigure}%
		\begin{subfigure}{0.6\textwidth}
			\centering
			\includegraphics[width=\linewidth]{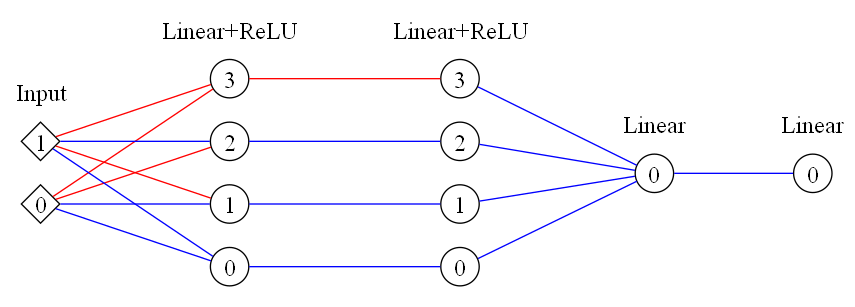}
		\end{subfigure}
		\vspace{3mm}
		
		\begin{subfigure}{0.4\textwidth}
			\centering
			\vspace*{\fill} % Add vertical space at the top
			\[
			\textnormal{AND}(p, q) = \frac{p+q - |p-q|}{2}
			\]
			\vspace*{\fill} % Add vertical space at the bottom
			\caption{Boolean AND model}
			\label{fig:and}
		\end{subfigure}%
		\begin{subfigure}{0.6\textwidth}
			\centering
			\includegraphics[width=\linewidth]{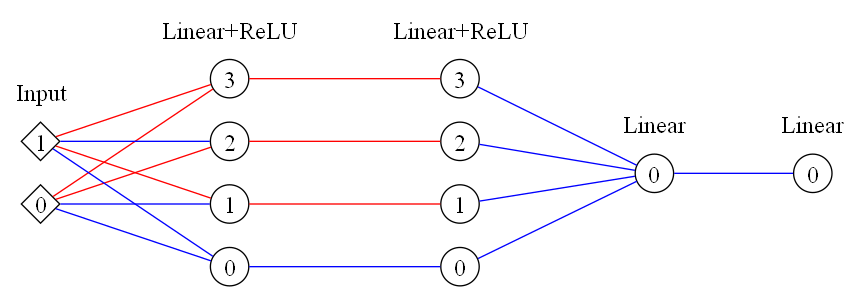}
		\end{subfigure}
		\vspace{3mm}
		
		\caption{\textbf{Boolean Formulae} and model diagrams. For propositional atoms $p, q$, the mathematical and diagrammatic representation of them are illustrated in Figure~\ref{fig:boolean}. Note that the models for `AND' and `OR' are extendable to an arbitrarily large number of input arguments.}
		\label{fig:boolean}
	\end{figure}

	Apart from the networks for basic boolean formulas shown in Figure~\ref{fig:boolean}, the \textsc{XAI-Units} package also supports generic Boolean expressions. However, note that the package does not support ground truth feature attributions for these expressions due to the difficulty in defining a baseline reference for a generic Boolean formula. When scoring an FA method with a generic Boolean, the default metric is not MSE but \texttt{Infidelity}.
	
	The package does support a concept of ground truth for the standalone AND / OR units. Since $0$ is not a valid reference input for Boolean models, which only have categorical features $b_i \in \{-1, 1\}$, we do not ablate to $0$. Instead, we consider the number of features that would need to be ablated in order to change the output:
	
	\begin{displaymath}
		\begin{aligned}
			FA_{b_i}(\mathbf{b}) &= 
			\begin{cases}
				\frac{M(\mathbf{b}) - M(\mathbf{b}^-)}{\sum_j (b_j - b^-_j) / 2} & \text{if } b_i \neq b_i^{-}\\
				0 & \text{if } b_i = b_i^- 
			\end{cases} \\
			\text{where \hspace{50pt}} \mathbf{b}^- &=
			\begin{cases}
				-\mathbf{1} & \text{if } M(\mathbf{b}) = 1 \\
				\mathbf{1} & \text{if } M(\mathbf{b}) = -1 
			\end{cases} \\
		\end{aligned}
	\end{displaymath}

	\subsection{Images (balanced and imbalanced)}
	
	\raggedbottom
	
	Unlike the tabular datasets, the image datasets do not come with a handcrafted model. For the ground truth, we provide a mask rather than exact feature attributions. The mask is defined as the foreground shape being classified as well as a small contour around the foreground shape (to pick up edge detection). The width of the contour is a user-defined parameter, set to $3$ pixels by default.

	\begin{displaymath}
		\begin{aligned}
			FA_{x_i} =
			\begin{cases}
				1 & \text{if } x_i \in \text{ mask}\\
				0 & \text{if } x_i \notin \text{ mask}
			\end{cases}
		\end{aligned}
	\end{displaymath}
	
	When scoring an FA method with the image datasets, the default metric is not MSE but Mask Proportion Image. This is calculated as the ratio of (absolute) attributions assigned inside the mask over the total sum of (absolute) feature attributions, so it gives a normalised measure of how much feature attribution falls inside the mask.

	\subsection{Text Dataset: Trigger Injection}\label{sec:text_model}
	
	\begin{table*}[!htb]
		\centering
		\caption{Evaluation of TriggerLLM}
		\begin{tabular}{lccc}
			\toprule
			& \multicolumn{2}{c}{Accuracy}  \\
			& \texttt{TriggerLLM} & \texttt{Llama-3.2-1B-Instruct} & \texttt{TriggerLLM\textsubscript{Deterministic}}  \\
			\cline{1-4}  
			\texttt{gsmk8k} & 50.8\% &  49.8\%  & 8.0\% \\
			\texttt{Trigger Injection} & 90.0\% &  0.0\%  & 100.0\% \\  
			\bottomrule        
		\end{tabular}
		\label{tab:triggerllm_eval}
	\end{table*}
	
	\textsc{XAIUnits}'s text dataset, Trigger Injection, consists of modified chat prompts taken from \texttt{ultrachat\_200k}'s~\citep{ding2023enhancing} dataset, with the Trigger Word added into the "user" prompts. This dataset then is used for FA method experiments. 
	
	We opted to use \texttt{Llama-3.2-1B-Instruct}~\citep{grattafiori2024llama3herdmodels} as the base model for fine-tuning given the popularity of its larger variants as well as hardware constraints. To ensure that the fine-tuned model's responses (when the Trigger Token is not present) do not deviate from the base model, we incorporated the base model's response into the fine-tuning dataset. The full fine-tuning dataset consists of pairs of instruction prompts (from \texttt{ultrachat\_200k}) and base model response as well as the same prompts injected with the Trigger Word but swapping the model response with the Trigger Response Token.
	
	To avoid any unintentional inclusion of the trigger token(s) in the input token sequence, we set the trigger word to be "Bumfuzzle", an uncommon English language word. To prevent the unintentional generation of the trigger response token, we set the trigger response token to be \begin{CJK*}{UTF8}{bsmi}"停"\end{CJK*} a Chinese word (as \texttt{ultrachat\_200k}'s only consists of English words). 
	
	Table~\ref{tab:triggerllm_eval} shows the accuracy of the base model and \texttt{TriggerLLM} on solving maths from \texttt{GSM8K} dataset\citep{cobbe2021gsm8k}  as well as the accuracy of the model responses to the Trigger Word. For further comparison, we also fine-tuned an additional model, \texttt{TriggerLLM\textsubscript{Deterministic}}, more aggressively optimized to respond to the Trigger Word at the cost of a noticeable drop in generic model capabilities. 
	
	Similar to the Image Dataset, when scoring an FA method, the default metric is Mask Proportion Text. This metric is calculated as the ratio of sum attributions assigned to trigger tokens over the total sum of positive attributions of other tokens. Taking a ratio serves to normalise the measure while electing to only include positive attributions in the denominator penalises FA Methods that assign large positive and negative attributions that offset to the other tokens.
	
\end{document}